%% file: main.tex
\documentclass[10pt,twocolumn,letterpaper]{article}

\usepackage{iccv}              

\definecolor{iccvblue}{rgb}{0.21,0.49,0.74}
\usepackage[pagebackref,breaklinks,colorlinks,allcolors=iccvblue]{hyperref}

\def\paperID{3552} 
\def\confName{ICCV}
\def\confYear{2025}

\usepackage{pifont}
\usepackage{makecell}

\newcommand{\modelname}{INP-CC }

\title{Open-Vocabulary HOI Detection with\\Interaction-aware Prompt and Concept Calibration}

\author{Ting Lei$^1$\quad Shaofeng Yin$^1$\quad Qingchao Chen$^2$\quad Yuxin Peng$^1$\quad Yang Liu$^{1}$\thanks{Corresponding author} \\
$^1$Wangxuan Institute of Computer Technology, Peking University \\
$^2$National Institute of Health Data Science, Peking University \\
{\tt\small \{ting\_lei, qingchao.chen, pengyuxin, yangliu\}@pku.edu.cn} \quad
{\tt\small yin\_shaofeng@stu.pku.edu.cn}
}

\begin{document}

\maketitle
\input{sec/0_abstract}
\input{sec/1_intro_v4}

\input{sec/2_related_work}

\input{sec/3_method}

\input{sec/4_experiments}

\input{sec/5_conclusion}

{
    \small
    \bibliographystyle{ieeenat_fullname}
    \bibliography{main}
}

\end{document}

%% file: sec/0_abstract.tex
\begin{abstract}

Open Vocabulary Human-Object Interaction (HOI) detection aims to detect interactions between humans and objects while generalizing to novel interaction classes beyond the training set.
Current methods often rely on Vision and Language Models (VLMs) but face challenges due to suboptimal image encoders, as image-level pre-training does not align well with the fine-grained region-level interaction detection required for HOI. Additionally, effectively encoding textual descriptions of visual appearances remains difficult, limiting the model’s ability to capture detailed HOI relationships.
To address these issues, we propose INteraction-aware 
Prompting with Concept Calibration (INP-CC), an end-to-end open-vocabulary HOI detector that integrates interaction-aware prompts and concept calibration. 
Specifically, we propose an interaction-aware prompt generator that dynamically generates a compact set of prompts based on the input scene, enabling selective sharing among similar interactions. This approach directs the model’s attention to key interaction patterns rather than generic image-level semantics, enhancing HOI detection.
Furthermore, we refine HOI concept representations through language model-guided calibration, which helps distinguish diverse HOI concepts by investigating visual similarities across categories.
A negative sampling strategy is also employed to improve inter-modal similarity modeling, enabling the model to better differentiate visually similar but semantically distinct actions.
Extensive experimental results demonstrate that INP-CC significantly outperforms state-of-the-art models on the SWIG-HOI and HICO-DET datasets. Code is available at https://github.com/ltttpku/INP-CC.

\end{abstract}

%% file: sec/1_intro_v4.tex
\section{Introduction}
\label{sec:intro}

\begin{figure}[t]
  \centering
   \begin{subfigure}{0.8\linewidth}
        \includegraphics[width=0.95\linewidth]{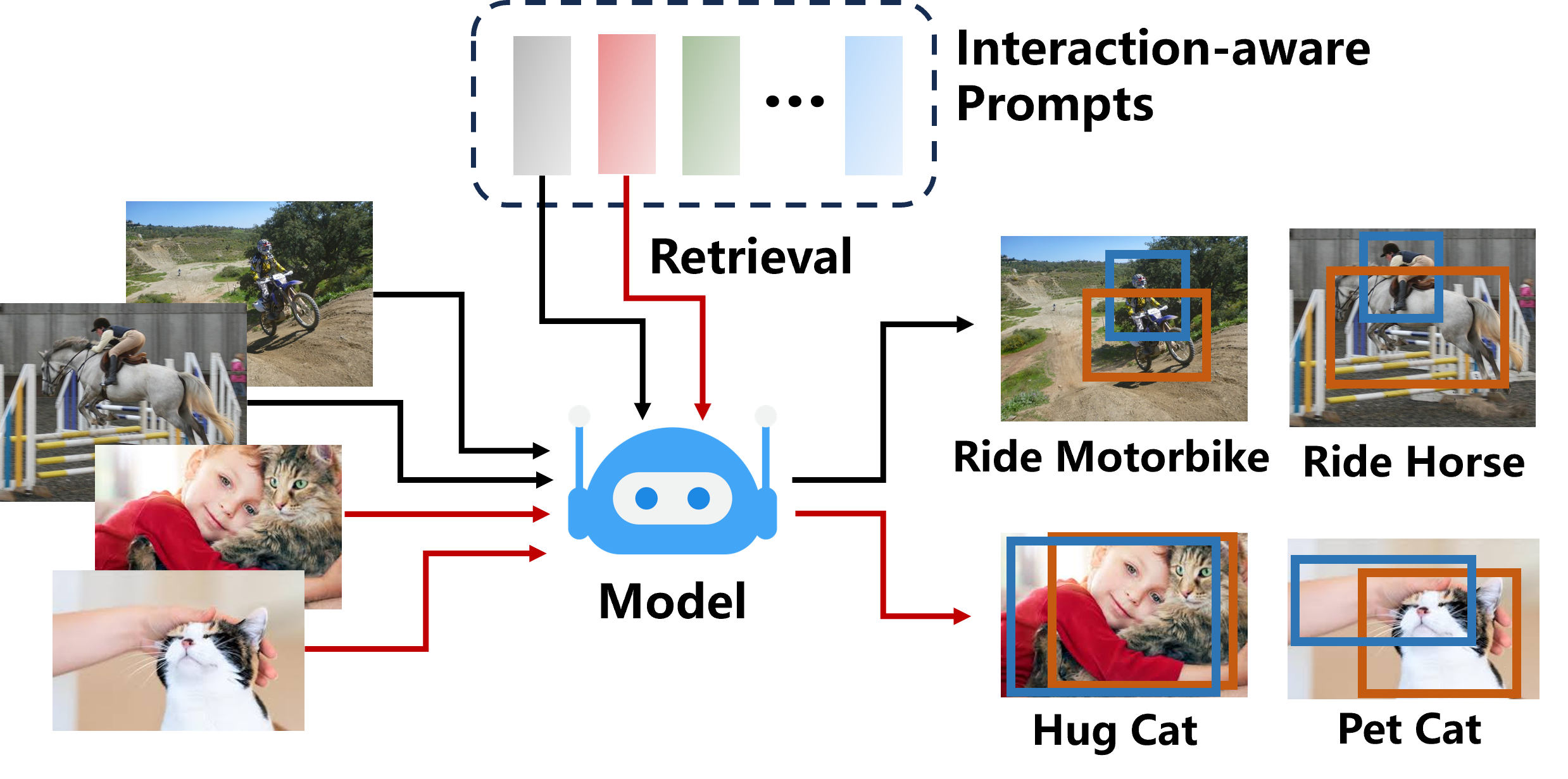}
        \caption{Interaction-aware prompt integration.}
        \label{fig:teaser_inp}
   \end{subfigure}
     
   \hfill
   \begin{subfigure}{0.9\linewidth}
        \includegraphics[width=0.48\linewidth]{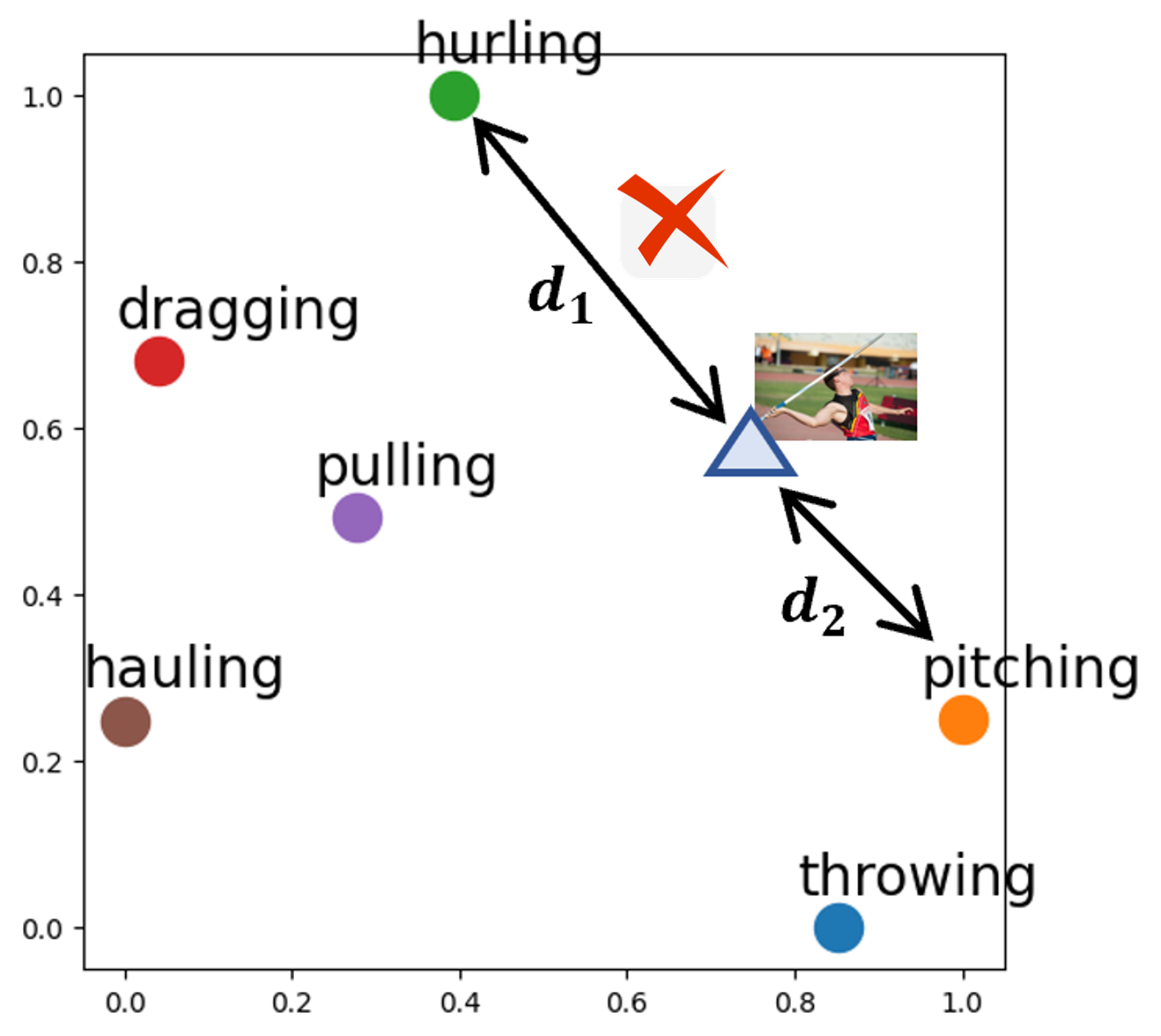}
        \includegraphics[width=0.48\linewidth]{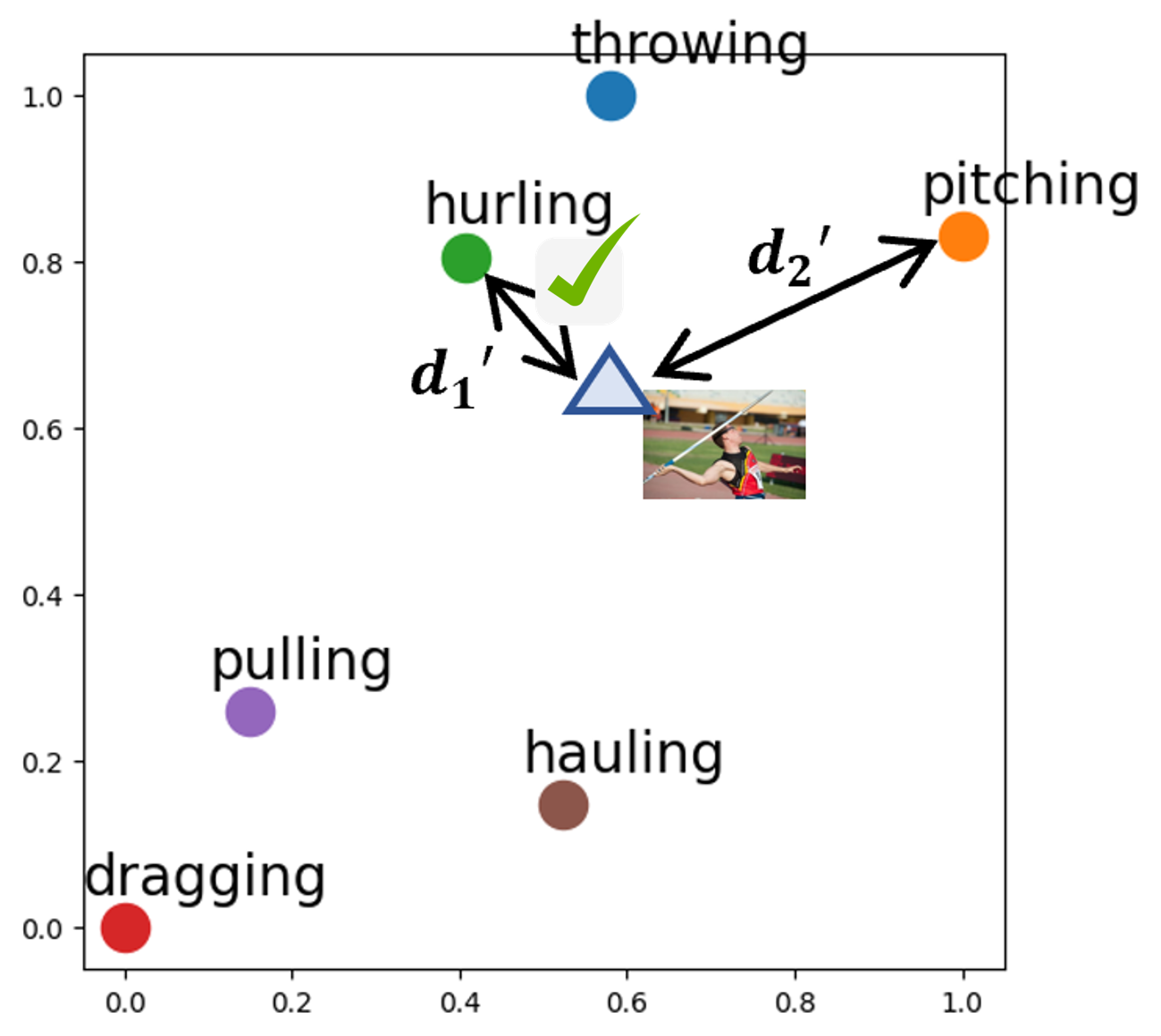}
        \caption{Intra- and inter-modal similarities in original CLIP space~(left) and our calibrated space~(right).}
        \label{fig:embed_comparison}
   \end{subfigure}
   \caption{(a) CLIP struggles with local region understanding due to its holistic pretraining. To address this, we introduce Interaction-Aware Prompts, enabling selective prompt sharing among related interaction categories. Similar interactions (e.g., hug and pet a cat) leverage the same prompt, improving region-level interaction detection. (b) Existing CLIP-based methods often struggle with adapting to diverse HOI concepts. For instance, the visual embedding of the action hurling (represented by the triangle) may misalign with the text embedding of pitching (represented by the orange circle). In contrast, our calibrated embeddings better distinguish visually similar actions, effectively capturing nuanced intra- and inter-modal correlations. }
   \label{fig:teaser}
   \vspace{-2em}
\end{figure}

Human-Object Interaction (HOI) detection involves identifying and localizing humans and objects within an image, while predicting their semantic relationships. Accurate HOI localization is crucial for applications such as video analysis~\cite{feichtenhofer2017spatiotemporal,yang2025planllm,zheng2024training,zheng2023generating}, human activity recognition~\cite{caba2015activitynet}, and scene understanding~\cite{li2024pixels,wang2024oed,yang2024active,gao2025conmo,mo2024bridging}. Deep neural networks~\cite{tian2023transformer,zhang2024deep,zhang2023fine,xu2024semantic,yang20243d} have demonstrated strong performance in HOI detection when provided with sufficient annotated training data~\cite{kim2021hotr,li2022improving,dong2022category,liu2022interactiveness_field,Zhang_2022_STIP,zhang2021CDN,zhang2023pvic,Xie_2023_CVPR,wang2024cyclehoi,li2024diffusionHOI}. However, these models primarily rely on closed-set training data, limiting their generalization to novel classes in open-world scenarios. In real-world settings, the presence of novel classes and the high cost of acquiring large annotated datasets for open-label tasks make it essential to develop transferable HOI detectors that can identify interactions based solely on class names in an Open Vocabulary (OV) setting.


Earlier works~\cite{kato2018compositional,bansal2020func,hou2020VCL,hou2021FCL,hou2021ATL} sought to enhance HOI detection by decomposing interactions into actions and objects, using data augmentation to create new combinations for detecting unseen interactions. However, without semantic context, these methods are limited to a small set of predefined actions and objects, restricting their generalization ability.
The development of visual-language models (VLMs) like CLIP~\cite{CLIP} has enabled recent research~\cite{wang2022_THID,ning2023hoiclip,liao2022gen,lei2024CMD-SE,wu2022EoID,guo2024HOIGen} to incorporate natural language into HOI detection, transferring knowledge from CLIP to identify previously unseen HOI concepts. 
Despite this progress, these approaches face key limitations in tackling open-vocabulary HOI detection: 
(1) Region vs. Image-Level Perception Misalignment: HOI detection requires fine-grained reasoning on localized image regions, whereas CLIP is trained on holistic image understanding. This discrepancy introduces a distribution gap, impeding the model’s detection performance in complex scenes.
(2) Suboptimal Text-Visual Alignment for Fine-Grained HOI Concepts: Current methods struggle to capture the subtle intra- and inter-modal relationships necessary for differentiating visually similar actions. This issue stems from the region-level misalignment mentioned above—CLIP’s image-level pretraining fails to establish fine-grained visual-text correspondences, leading to the loss of critical action distinctions. For instance, words like “hurl”, “throw”, and “pitch” have distinct meanings but are often weakly aligned with visual cues. Similar backgrounds or irrelevant regions further disrupt learning, making it hard to distinguish actions like “hurling” vs. “pitching” (see~\cref{fig:embed_comparison}), limiting adaptability to novel interactions.


\begin{figure}[t]
  \centering
   \begin{subfigure}{0.3\linewidth}
        \includegraphics[width=0.95\linewidth]{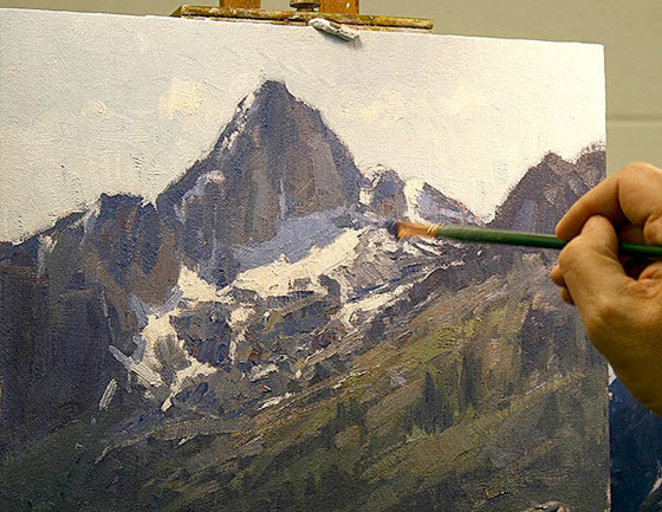}
        \caption{Origin image.}
        \label{fig:image_vs_region_a}
   \end{subfigure}
   \begin{subfigure}{0.3\linewidth}
        \includegraphics[width=0.95\linewidth]{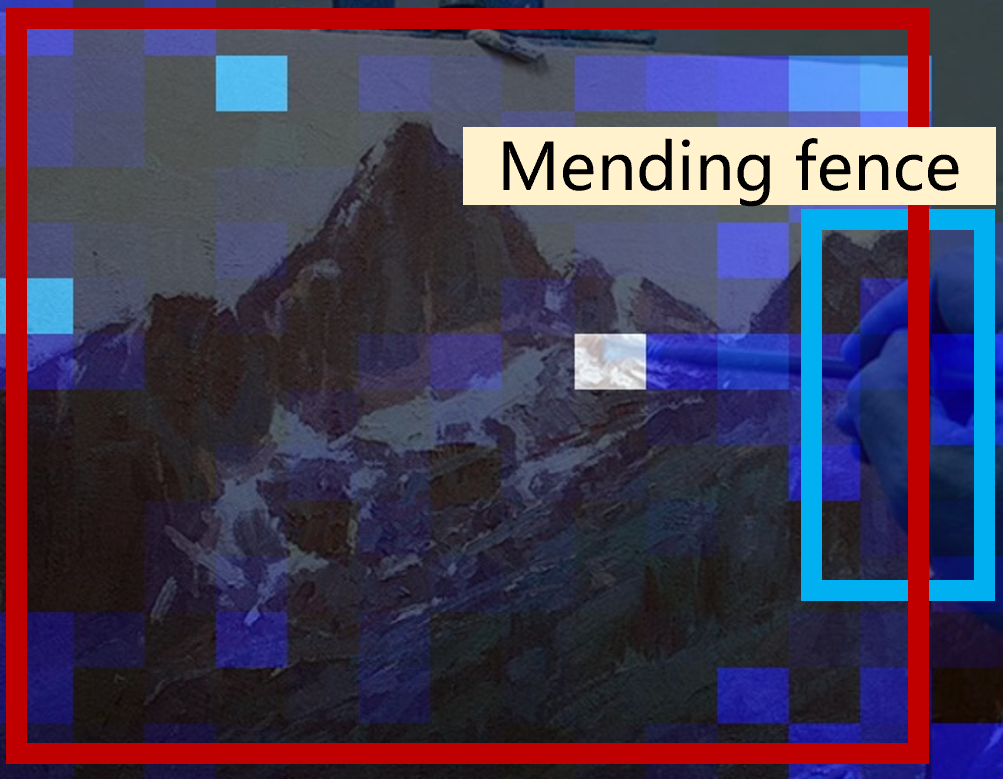}
        \caption{Previous method.}
        \label{fig:image_vs_region_b}
   \end{subfigure}
   \begin{subfigure}{0.3\linewidth}
        \includegraphics[width=0.95\linewidth]{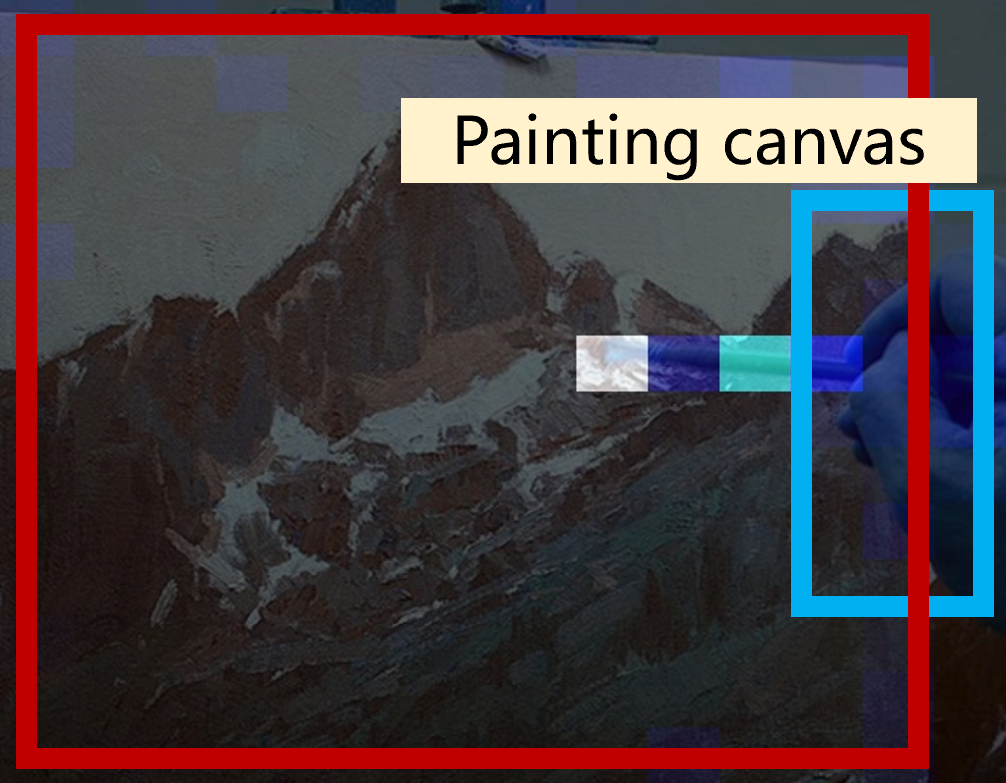}
        \caption{Ours.}
        \label{fig:image_vs_region_c}
   \end{subfigure}
   \caption{Qualitative comparison between previous and our method. The previous method (CMD-SE) struggles to accurately identify key interaction regions, leading to incorrect predictions (such as ``mending fence"). In contrast, our method specifically focuses on the interaction between the paintbrush and the canvas, enabling more accurate predictions of ``painting canvas".}
   \label{fig:compare}
   \vspace{-1.5em}
\end{figure}

To address the challenges mentioned, we introduce INP-CC, a HOI generalist model with INteraction-aware Prompt and Concept Calibration, designed for HOI detection in open-world settings. The overall framework enables the model to focus attention on key interaction regions, as demonstrated in~\cref{fig:compare}.

\textit{First}, to mitigate the perception gap in the pretrained image encoder, we propose interaction-aware prompts that guide the model’s focus toward interaction-relevant regions. Specifically, we develop an interaction-adaptive prompt generator, which constructs a compact set of interaction-specific prompts. These prompts consist of (1) a common prompt, shared across all scenes to capture general HOI features, and (2) interaction prompts, selectively shared among semantically related interactions. 
Instead of assigning a unique prompt to each interaction—which is infeasible in an open-vocabulary setting due to the vast interaction space and high computational cost—we enable selective sharing of prompts among interactions with similar semantic or functional patterns, as illustrated in~\cref{fig:teaser_inp}. For example, interactions such as ``hold cup" and ``hold bottle" exhibit similar hand-object contact dynamics, making a shared prompt beneficial for efficient representation learning. 
To further enhance adaptability, our approach dynamically selects and refines these prompts based on input images, enabling better contextualization of region-level interactions. 

\textit{Second}, we propose calibrating HOI concept representations with guidance from large language models (LLMs). We leverage GPT-3.5 to generate detailed visual descriptions for each HOI category, covering human poses, object attributes, and interaction contexts. However, CLIP struggles with aligning region-level visual features with text, favoring dominant scene elements over fine-grained interactions~\cite{yuksekgonul2022and,yamada2022lemons,kim2023region}. Consequently, text embeddings may not align well with the interactive components in the image.
To address this, we incorporate Instructor Embedding~\cite{INSTRUCTOR}, a T5-based discriminative language model, to refine and reassess category similarities. This improves intra-modal relationships, clustering related actions like “hurling” and “throwing” more effectively, as shown in~\cref{fig:embed_comparison}.
Additionally, we introduce a negative sampling strategy, selecting hard negatives from semantically similar concepts, ensuring the model learns to distinguish fine-grained action differences. As illustrated in~\cref{fig:embed_comparison}, this strategy pulls the visual embedding of ``hurling" closer to its corresponding text embedding while pushing it away from the text embedding of ``pitching", reducing confusion between visually similar but semantically distinct actions.

The main contributions of this paper are threefold. 
(1) We introduce an interaction-adaptive prompt generator, a novel mechanism for prompt generation and selection that enhances the model’s adaptability to various scenes by incorporating interaction-specific knowledge.
(2) We leverage large language models to capture HOI concept relationships, using this information to sample negative categories for enhancing inter-modal modeling.
(3) Experiments on SWIG-HOI~\cite{wang2021SWIG-HOI} and HICO-DET~\cite{chao2018HICO-DET} datasets demonstrate that our proposed method achieves state-of-the-art performance in Open-Vocabulary HOI detection.

%% file: sec/2_related_work.tex
\section{Related Work}
\label{sec:related_work}

\subsection{HOI Detection}
Based on network architecture design, prior HOI detection methods can be categorized into two main paradigms: one-stage~\cite{zhong2021glance,liao2020ppdm,fang2020dirv,chen2021reformulating,kim2020uniondet,gkioxari2018detecting,kim2022mstr,zhang2021CDN,Tu_2023_ICCV,Kim_2023_CVPR,li2023sqab,wang2022chairs,zhou2022human,cheng2024parallel,chen2023qahoi,iftekhar2022look,li2024diffusionHOI} and two-stage~\cite{xu2019learning,cao2023RmLR,zhang2023pvic,zhang2022UPT,gao2020drg,li2019transferable,gao2018ican,Park_2023_CVPR,ting2023hoi,wang2024bilateral,jiang2024SCTC,wu2024pose_aware,wang2022distance} approaches.
In the two-stage approach, an off-the-shelf detector~\cite{ren2016faster_rcnn,carion2020DETR} first locates and classifies objects, followed by dedicated modules for associating humans with objects and recognizing their interactions. These methods often utilize multi-stream~\cite{gupta2019no,li2020pastanet,liu2022interactiveness_field,hou2021FCL} or graph-based~\cite{ulutan2020vsgnet,yang2020graph} techniques to improve interaction understanding. 
In contrast, one-stage methods apply multitask learning to simultaneously handle instance detection and interaction recognition~\cite{liao2020ppdm,zhang2021CDN,tamura2021qpic,kim2021hotr}. Despite these advancements, traditional HOI detection approaches treat interactions as discrete labels within a fixed category space, relying on classifiers trained on predefined categories. This constraint limits their capacity to recognize the broader spectrum of potential unseen interactions.

\subsection{VLM-based HOI Detection}
Despite substantial progress in HOI detection, most methods still treat interactions as discrete labels, often missing the richer semantic structure embedded in triplet labels. 
Earlier research~\cite{zhong2020polysemy,iftekhar2022SSRT,zhou2019RLIP,Lei_2024_eccv} tackled this problem by integrating language-based priors, which enhanced the model’s ability to generalize to unseen HOI triplets during inference.
With the rise of Vision-Language Models (VLMs) like CLIP, recent studies~\cite{wan2024exploiting,wan2023weaklyHOI,xue2024KI2HOI,liao2022gen,ning2023hoiclip,li2024logichoi,mao2023clip4hoi,yang2024CaCLIP,wu2022EoID,guo2024HOIGen,lei2024CMD-SE,yang2024MPHOI} have shifted focus toward transferring VLM knowledge to recognize novel HOI concepts. For example, GEN-VLKT~\cite{liao2022gen} and HOICLIP~\cite{ning2023hoiclip} utilize the CLIP visual encoder to guide interaction representation learning, initializing classifiers with CLIP-generated text embeddings. OpenCat~\cite{Zheng_2023_CVPR} utilizes weakly supervised data and introduces various proxy tasks to pre-train HOI representations with CLIP. Additionally, RLIP~\cite{zhou2019RLIP} and DP-HOI~\cite{li2024disentangledPretrain} explore vision-language pretraining to develop transferable HOI detectors. THID~\cite{wang2022_THID} introduces a HOI sequence parser capable of identifying multiple interactions simultaneously, demonstrating strong performance on the open-vocabulary SWIG-HOI dataset~\cite{wang2021SWIG-HOI}. MP-HOI~\cite{yang2024MPHOI} integrates visual prompts into existing language-guided-only HOI detectors based on large-scale training. CMD-SE~\cite{lei2024CMD-SE} further enhances HOI detection by incorporating conditional multi-level decoding along with the human body part descriptions to enrich interaction context. 
However, previous VLM-based methods still fail to accurately model the diversity of HOI concepts due to a mismatch between the region-level visual representations and the holistic, general-purpose nature of the text-based representations.


\subsection{Large Language Model}
Language data has become essential in open-vocabulary research, with Large Language Models (LLMs) offering valuable knowledge for various NLP tasks. Works such as~\cite{menon2022visual, pratt2023does, yang2023language} have leveraged LLMs to generate descriptive labels for visual categories, enriching VLMs without the need for additional training or labeling. Recent studies~\cite{unal2023weaklysupervised, li2023zeroshot, kaul2023multimodal, jin2024llms, zang2023contextual,du2025lami} further utilize LLMs to generate fine-grained descriptions for detection tasks. For instance, \cite{zang2023contextual} uses contextual LLM tokens as conditional queries for the visual decoder, and \cite{unal2023weaklysupervised} filters out unlikely HOIs using LLM semantic priors. Recent works employ \textit{generative} LLMs to generate fine-grained HOI descriptions. UniHOI~\cite{cao2023UniHOI} acquires HOI descriptions through knowledge retrieval.  SICHOI~\cite{luo2024SIC} constructs a syntactic interaction bank from multiple levels to provide interaction clues. CMD-SE~\cite{lei2024CMD-SE} exploits descriptions of body part states to discern fine-grained HOIs. 
However, previous methods still struggle to model intra- and inter-modal similarities across diverse HOI concepts. To address this, we propose enhancing VLMs with \textit{discriminative} LLMs to capture intra-modal relationships and distinguish visually similar yet semantically distinct actions through a hard negative sampling strategy for inter-modal modeling. 


%% file: sec/3_method.tex
\section{Method}
\label{sec:method}

In this section, we propose the design of an open-vocabulary HOI detector capable of identifying interactions from novel classes by leveraging the generalization learned from a set of trained classes.  

\begin{figure*}[t]
  \centering
   \includegraphics[width=0.85\linewidth]{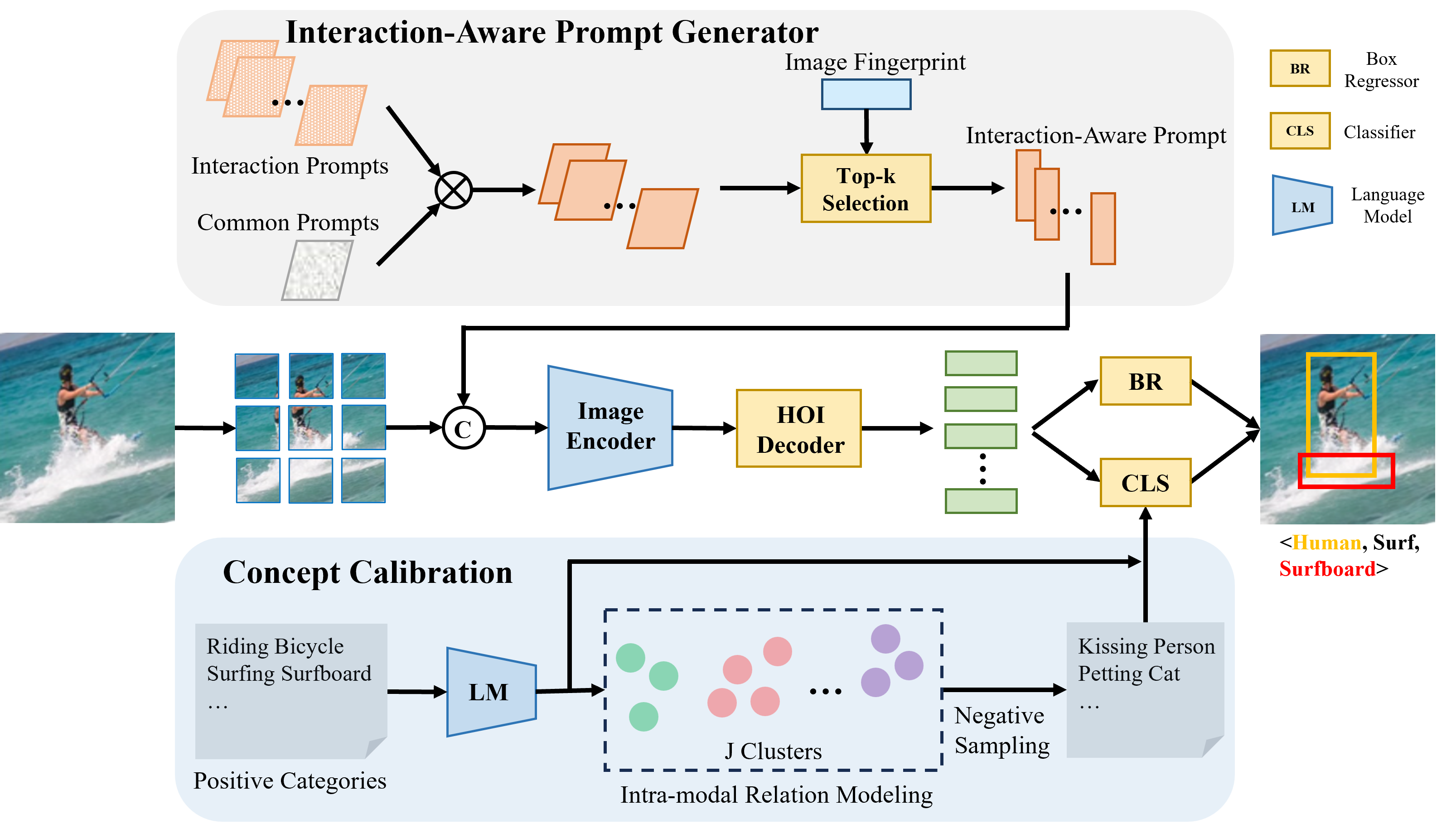}
   \caption{Overall architecture of our \modelname method. Given an input image, we first extract the spatial feature map using a pre-trained CLIP image encoder, augmented with our interaction-aware prompts. These prompts consist of a common prompt, shared across all scenes, and individualized prompts that are tailored to specific interaction patterns. We then adaptively select and refine these prompts based on the input image. To further enhance HOI recognition, we use embeddings calibrated via language models to model intra-modal relationships, producing K distinct clusters. During training, we sample negative categories from the clusters that overlap with the positive categories in the current batch, improving inter-modal similarity modeling. This approach helps create a more structured and discriminative semantic space for HOI concepts.}
   \label{fig:method}
   \vspace{-1.5em}
\end{figure*}

\subsection{Preliminary}
\label{subsec:preliminary}

\noindent \textbf{Problem Definition.} 
Open Vocabulary HOI Detection aims to detect HOIs from novel classes besides the base classes on which the detector is trained. We define an interaction as a tuple $(b_{o}, b_{h}, o, a)$, where $b_{o}, b_{h} \in \mathbb{R}^{4}$ denote the bounding box of a human and object instance, respectively. $o \in \mathbb{O}$ and $a \in \mathbb{A}$ denote the object and action category, where $\mathbb{A}$ and $\mathbb{O}$ denote the human action and object set, respectively.
In the training stage, we have HOIs consisting of base action classes $\mathbb{A}_\mathrm{B}$ and base object classes $\mathbb{O}_\mathrm{B}$. In the testing stage, there are novel HOI classes that involve either $\mathbb{A}_\mathrm{N}$ or $\mathbb{O}_\mathrm{N}$ that need to be detected. 
Typically, $\mathbb{A}_\mathrm{B} \cup \mathbb{A}_\mathrm{N} = \mathbb{A}$ and $\mathbb{A}_\mathrm{B} \cap \mathbb{A}_\mathrm{N} = \emptyset$; $\mathbb{O}_\mathrm{B} \cup \mathbb{O}_\mathrm{N} = \mathbb{O}$ and $\mathbb{O}_\mathrm{B} \cap \mathbb{O}_\mathrm{N} = \emptyset$.
In other words, the objective of open-vocabulary HOI detection entails recognizing interactions that have not been encountered during the training phase, encompassing unseen objects, actions, and their various combinations.

\noindent \textbf{Baseline Revisited.}
We begin by introducing a basic end-to-end open-vocabulary HOI detector, leveraging the generalization capability of CLIP~\cite{CLIP} on classification tasks. We frame HOI detection as an end-to-end set matching problem, similar to DETR~\cite{carion2020DETR}, thereby eliminating the need for handcrafted components such as anchor generation.
Firstly, for a given image $\mathbf{I}$, we obtain the spatial feature map $F_{\mathbf{I}}$ using the pre-trained CLIP image encoder $\mathrm{E_V}$:
\begin{equation}
  F_{\mathbf{I}} = \mathrm{E_V}(\mathbf{I})
  \label{eq:encode_image}
\end{equation}
where $F_{\mathbf{I}} \in \mathbb{R}^{HW\times C}$ denotes a sequence of feature embeddings of $\mathbf{I}$. We then utilize a transformer-based decoder $D$ to decode the HOIs in $\mathbf{I}$. Specifically, taking the projected context feature and HOI queries $Q = (q_1, q_2, ..., q_N)$ as input, where $N$ denotes the number of HOI queries, the output from the final layer of $\mathrm{D}$ serves as the representation of interactions:
\begin{equation}
    H = \mathrm{D}(Q, F_{\mathbf{I}})
\end{equation}
where $Q$ is treated as query, and the projected context representation is treated as key and value during the cross-attention mechanism of the HOI decoder $\mathrm{D}$.
The resulting $H = (h_1, h_2, ..., h_N)$ corresponds to the decoded interaction features associated with each query.
Then, we feed them to two different head networks: 1) a bounding box regressor $\mathrm{P_{bbox}}$, which predicts a confidence score $c$ and the bounding box of the interacting human and object $(b_h, b_o)$. 2) a linear projection $\mathrm{P_{cls}}$ which maps the feature to the joint vision-and-text space. Similar to~\cite{wang2022_THID, lei2024CMD-SE}, we compute its similarity with the semantic features $T_{hoi}$ from the CLIP text encoder $\mathrm{E_T}$ for interaction recognition.

\noindent \textbf{Overall Architecture.}
The overall architecture of our proposed \modelname is illustrated in~\cref{fig:method}.
Given an image, we obtain the spatial feature map $F_{\mathbf{I}}$ using the pre-trained CLIP image encoder $\mathrm{E_V}$ along with our proposed interaction-aware prompt $P_{IA}$, as shown in the upper part of~\cref{fig:method}. The interaction-aware prompt generation process, detailed in~\cref{subsec:interaction_aware_prompt}, bridges the gap between image-level recognition and region-level interaction detection, improving the encoder's ability to model diverse HOI scenes.
Then in~\cref{subsec:concept_calibration}, we describe the process of HOI concept calibration, including intra-modal relationship modeling and negative concept sampling strategy~(illustrated in the lower part of~\cref{fig:method}, shaded in light blue). This approach enhances the model's ability to discern correlations among diverse HOI concepts, guiding the model within a more structured semantic space.
At last, we present loss functions used to train our \modelname in~\cref{subsec:training}.

\subsection{Interaction-aware Prompt Generation}
\label{subsec:interaction_aware_prompt}

To bridge the gap between image-level pre-training and the detailed region-level interaction detection needed for HOI tasks, we introduce interaction-aware prompts to strengthen the visual encoder’s ability to focus on key interaction regions and represent diverse HOI scenes. In this subsection, we present an interaction-adaptive prompt generator designed to capture key attributes specific to HOIs, effectively transferring pre-trained VLM knowledge into the HOI context. These interaction prompts are then refined through an adaptive selection mechanism to optimize their relevance and impact.

\noindent \textbf{Constructing Interaction-aware Prompts.} 
As illustrated in the gray shaded area in~\cref{fig:method}, our interaction adaptive prompt generator introduces a targeted mechanism for generating prompts tailored to the HOI detection task. It comprises two key components: (1) a common prompt, which captures shared knowledge across all interaction classes, and (2) a set of interaction-specific prompts, efficiently encoded through low-rank decomposition to represent interaction-specific knowledge. To enhance both efficiency and generalizability, we constrain the number of interaction prompts to be significantly lower than the total number of interaction classes in the open-vocabulary setting, encouraging semantically similar interactions to share the same prompt. This enables each interaction prompt to adaptively capture commonalities across different HOI types.

Specifically, we define $P_{C} \in \mathbb{R}^{L \times D}$ as the common prompt, where $L$ is the prompt length and $D$ denotes the feature embedding dimension. This common prompt captures general knowledge applicable to all classes. Additionally, we construct a series of prompts tailored to specific interactions, denoted as $\{\hat{P}_{IT}^{i}\}_{i=1}^{M}$, where $M$ represnets the total number of interaction patterns, and each prompt $\hat{P}_{IT}^{i} \in \mathbb{R}^{L \times D}$ is dedicated to learning interaction-specific information. Each interaction-specific prompt $P_{IT}^{i}$ is obtained by combining the common prompt with the interaction-specific prompt as follows: \begin{equation} 
    P_{IT}^{i} = \hat{P}_{IT}^{i} \odot P_{C}
\end{equation} 
where $\odot$ denotes the Hadamard product.

To enhance computational efficiency and performance, inspired by effective low-rank methods~\cite{li2018measuring}, we parameterize $\hat{P}_{IT}^{i}$ as a low-rank matrix, represented by the product of two low-rank vectors: $u_{IT}^i \in \mathbb{R}^L$ and $v_{IT}^i \in \mathbb{R}^D$. This decomposition into rank-one subspaces facilitates the efficient encoding of interaction-specific information, enabling the model to leverage both common and interaction-specific knowledge effectively for precise HOI detection.

\noindent \textbf{Selection Mechanism.}
To enable adaptive learning of interaction prompts, we introduce a selection mechanism that dynamically chooses the appropriate interaction prompts for a given input image. Specifically, each interaction prompt is associated with a learnable key, defined as: 
\begin{equation} 
    k_{IT}^{i} = g(P_{IT}^{i}) 
\end{equation} 
where the function $g$ is implemented as a Multi-Layer Perceptron (MLP) layer. For an input image $\mathbf{I}$, we compute the similarity between its feature embedding and each interaction key: 
\begin{equation} 
    w_i = \phi(f_\mathbf{I}, k_{IT}^{i}) 
\end{equation} 
where $f_\mathbf{I}$ is the image fingerprint extracted by the CLIP image encoder $\mathrm{E_V}$, $\phi$ denotes the cosine similarity function, and $w_i$ is a weight vector that reflects the relevance of each prompt.
We then select the top-$k$ most relevant prompts based on their similarity weights $w_i$, forming a set of selected keys $\mathbb{K}_{IT}$. The final interaction-aware prompt is constructed by combining the selected prompts weighted by their respective similarities: 
\begin{equation}
    P_{IA} = \sum\limits_{k_{IT}^{i} \in \mathbb{K}_{IT}} w_i P_{IT}^{i}
\end{equation}
By selecting prompts in an input-adaptive manner, this mechanism balances the common and specific attributes of the HOI classes. Consequently, it enables the model to contextualize interactions more effectively, enhancing the pre-trained encoder's ability to capture distinctive HOI characteristics.
Note that our design also plays a crucial role in guiding the model to focus on interaction-relevant regions. When multiple semantically similar interactions (e.g., ``hold cup" and ``hold bottle")
share a common interaction prompt, the model learns to emphasize key regions that are consistently important across those interactions—such as the hand-object contact area. This reinforcement improves the model's ability to localize fine-grained interaction-specific details while filtering out irrelevant background noise. Additionally, by sharing prompts among semantically related interactions, the model benefits from richer supervision over common interaction attributes, making it easier to generalize and refine attention to relevant interaction regions, even in novel scenarios.

\begin{figure}[t]
  \centering
   \includegraphics[width=0.75\linewidth]{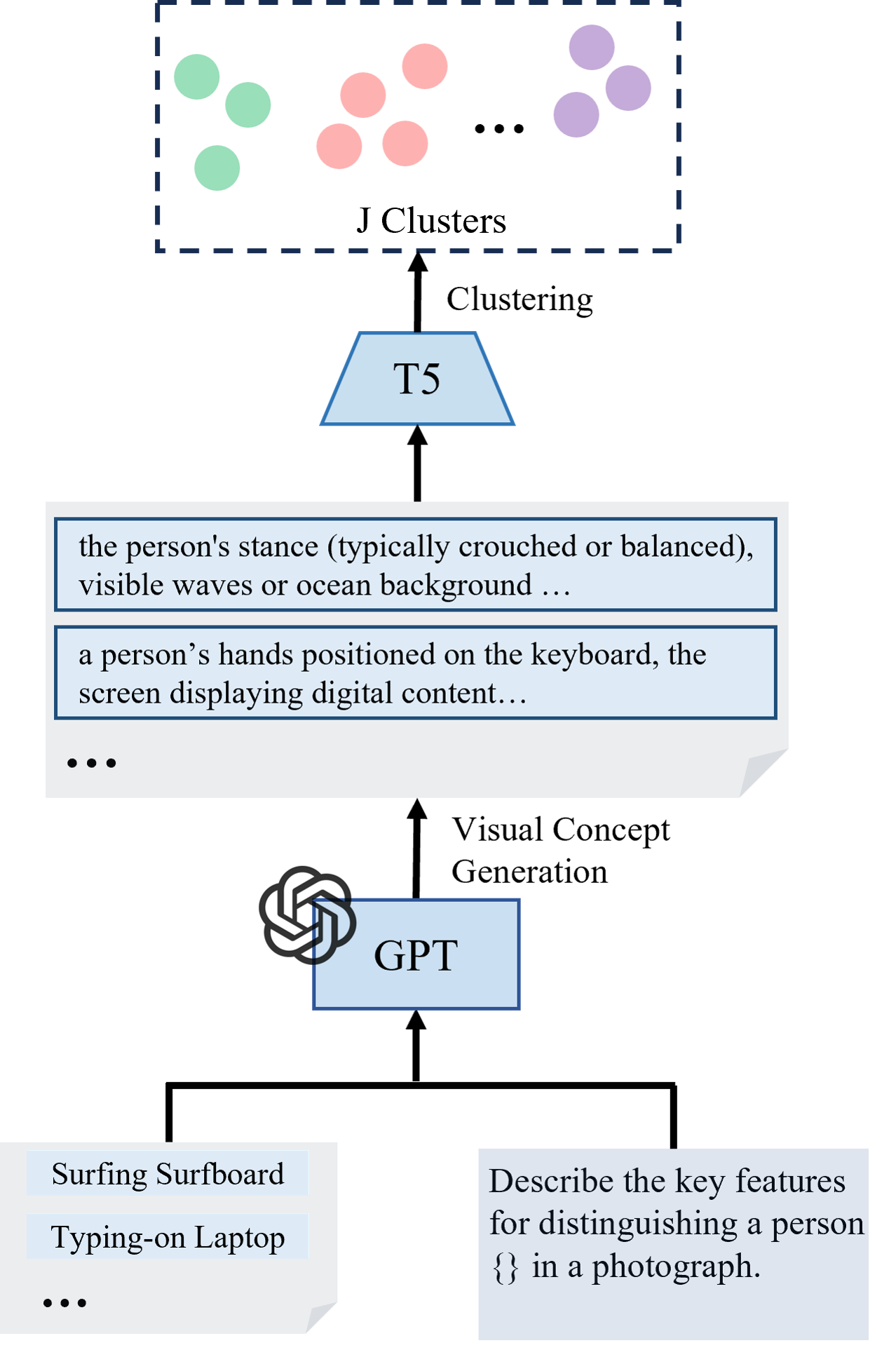}
   \caption{Intra-modal Relationship Modeling. }
   \label{fig:clustering}
   \vspace{-1em}
\end{figure}

\subsection{HOI Concept Calibration}
\label{subsec:concept_calibration}
To alleviate VLM's weak adaptability to diverse HOI concepts, we propose to leverage the language models to calibrate the semantic representations, enhancing the model's ability to recognize and cluster visually similar actions effectively.
In this subsection, we first introduce the process of intra-modal relationship modeling using language models, which enables more refined concept alignment. Additionally, we incorporate a negative sampling strategy during training to improve the model's capacity to differentiate between visually similar but conceptually distinct actions.

\noindent \textbf{Intra-modal Relation Modeling.}
Based on the problem identified in~\cref{fig:embed_comparison}, we generate visual descriptions for each concept using GPT, establishing clearer visual distinctions and enhancing concept representation. Additionally, we leverage the instruction embedding, provided by a T5 model~\cite{ni2021sentence} 
, known for its robust semantic understanding, to assess similarity relationships among these visual concepts, thereby capturing intra-modal relationships.
As illustrated in~\cref{fig:clustering}, for a given HOI category $(a, o)$—where $a \in \mathbb{A}$ and $o \in \mathbb{O}$ represent an action and an object, respectively, as defined in~\cref{subsec:preliminary}—we begin by generating fine-grained visual descriptors $d_{a,o}$ by prompting GPT:
\begin{equation}
    d_{a,o} = \mathrm{GPT}(a, o, \text{prompt})
\end{equation}
These visual descriptions are then processed by the T5 model $\mathrm{E_{LM}}$ to generate visual description embeddings:
\begin{equation}
    T_{vis}^{a,o} = \mathrm{E_{LM}}(d_{a,o})
\end{equation}
This approach constructs an open set $\mathbb{D}$ of visual concepts and their corresponding embeddings $T_{vis}$. To identify visually similar concepts, we cluster the visual description embeddings into $J$ clusters, denoted as $\mathbb{C} = \{C_i\}_1^J$. Concepts within the same cluster are considered to share similar visual attributes. As shown in~\cref{fig:method}, these extracted intra-modal relationships are then utilized during training in the negative concept sampling strategy, which reinforces the model's ability to discern visually comparable yet semantically distinct actions.

\noindent \textbf{Negative Category Sampling.}
To improve inter-modal similarity modeling, we propose a negative category sampling strategy. In this approach, negative samples are drawn from semantically similar concepts within the current minibatch during classification loss computation. This helps the model differentiate between visually similar yet semantically distinct actions more effectively.

Specifically, to ensure the model remains focused on visually relevant distinctions, we sample negative categories from clusters that include the ground truth categories of each training batch, denoted as $\mathbb{C}_{cur}$. This guarantees that the model is exposed to visually similar yet distinct categories, forcing it to differentiate subtle differences between related HOI concepts.
As a result, by sampling negative categories from these clusters, we encourage the visual embeddings of HOIs to align more closely with their corresponding text embeddings, while simultaneously distancing them from the embeddings of similar, neighboring categories. For instance, as illustrated in~\cref{fig:embed_comparison}, the visual embedding of ``hurling” is encouraged to align more closely with its corresponding text embedding, rather than being confused with the embedding of ``pitching”. This strategy helps the model learn finer distinctions between visually similar categories, thereby enhancing its generalization ability and improving robustness.

\subsection{Training and Inference}
\label{subsec:training}

\noindent \textbf{Training.}
We compute the box regression loss $\mathcal{L}_{b}$, the intersection-over-union loss $\mathcal{L}_{iou}$, and the interaction classification loss $\mathcal{L}_{cls}$ during training following previous work~\cite{liao2022gen,ning2023hoiclip,lei2024CMD-SE}:
\begin{equation}
    \mathcal{L} = \lambda_b \sum\limits_{i\in\{h,o\}} \mathcal{L}_{b}^{i} \ 
    + \lambda_{iou} \sum\limits_{i\in\{h,o\}} \mathcal{L}_{iou}^{i} \ 
    + \lambda_{cls} \mathcal{L}_{cls}
\label{eq:loss}
\end{equation}

\noindent \textbf{Inference.}
For each HOI prediction, including the bounding box pair $(\hat{b_{h}^{i}}, \hat{b_{o}^{i}})$, the bounding box score $\hat{c_{i}}$ from the bounding box regressor, and the interaction score $\hat{s_{i}}$ from the interaction classifier, the final score $\hat{s_{i}}'$ is computed as:
\begin{equation}
    \hat{s_{i}}' = \hat{s_{i}} \cdot \hat{c_{i}}^{\gamma}
\end{equation}
where $\gamma$ is a constant used during inference to suppress
overconfident objects~\cite{zhang2021scg,zhang2022UPT}.

%% file: sec/4_experiments.tex
\section{Experiment}
\label{sec:experiment}

\subsection{Experimental Setting}
\label{subsec:exp_setting}

\noindent \textbf{Datasets.}
Our experiments are mainly conducted on two datasets, SWIG-HOI~\cite{wang2021SWIG-HOI} and HICO-DET~\cite{chao2018HICO-DET}.
The SWIG-HOI dataset encompasses more than 400 human actions and 1000 object categories. Notably, the test set of SWIG-HOI naturally contains previously unseen combinations due to the extensive category space of actions and objects. 
The HICO-DET dataset provides 600 combinations involving 117 human actions and 80 objects. We follow~\cite{hou2020VCL,wang2022_THID} to simulate a zero-shot detection setting by holding out 120 rare interactions from all 600 interactions.

\noindent \textbf{Evaluation Metric.}
We follow the settings of previous works~\cite{chao2018HICO-DET,liu2022interactiveness_field,liao2022gen,wang2022_THID} to use the mean Average Precision (mAP) for evaluation. We define an HOI triplet prediction as a true-positive example if the following criteria are met: 1) The IoU of the human bounding box and object bounding box are larger than 0.5 \textit{w.r.t.} the GT bounding boxes; 2) the predicted interaction category is accurate.

\noindent \textbf{Implementation Details.}
We employ the ViT-B/16 version as our visual encoder following~\cite{wang2022_THID,lei2024CMD-SE}. We set the size of the interaction-aware prompt to 128 and 8 for SWIG-HOI and HICO-DET datasets, respectively. We set $k$ in the selection mechanism to 2. We set the cluster number $J$ to 64 and select 10 hard negative samples during each iteration. We train our model for 80 epochs with a batch size of 128 on 2 NVIDIA 3090 GPUs. 

\subsection{Comparison with Other Methods}
\label{subsec:compare_sota}

\noindent \textbf{Results on HICO-DET.}
We compare our method with state-of-the-art approaches in the zero-shot setting of HICO-DET (\cref{tab:hico-det}). While recent methods~\cite{liao2022gen,ning2023hoiclip} use CLIP text embeddings for interaction classification, they rely on DETR models pretrained on COCO~\cite{lin2014COCO}, limiting scalability in open-world scenarios. However, this comparison is not entirely fair, as HICO-DET and COCO share the same object labels. Open-vocabulary methods~\cite{wang2022_THID,lei2024CMD-SE} overcome this constraint by avoiding detection dataset pretraining. Our \modelname outperforms CMD-SE by 0.77\% 
in mAP on the full split, setting a new state-of-the-art.

\noindent \textbf{Results on SWIG-HOI.}
To evaluate the effectiveness of our approach, we compare the results with previous state-of-the-art methods on the SWIG-HOI dataset.
As shown in~\cref{tab:swig-hoi}, our model significantly outperforms the previous state-of-the-art~(CMD-SE) on all splits, achieving a relative gain of 9.70\%
on all interactions. This shows the strong capability of our \modelname to detect and recognize the interactions of human-object pairs in the open-vocabulary scenario. 
Additionally, our model surpasses CMD-SE by 1.48\%, 1.35\%, and 1.59\% on the interactions ``admiring", ``browsing", and ``reading", respectively, effectively distinguishing fine-grained variations in ``looking" behaviors.

\begin{table}
  \centering
  \begin{tabular}{@{}lcccc@{}}
    \toprule
    Method & {\makecell[c]{Pretrained \\ Detector}} & Unseen & Seen & Full \\
    \midrule
    \multicolumn{4}{@{}l@{}}{\textit{Zero-shot Methods}} \\
    \hline
    GEN-VLKT~\cite{liao2022gen}  &  \ding{51}  & 21.36 & 32.91 & 30.56 \\
    HOICLIP~\cite{ning2023hoiclip}  & \ding{51} & 23.48 & 34.47 & 32.26 \\
    CLIP4HOI~\cite{mao2023clip4hoi} & \ding{51} & 28.47 & \textbf{35.48} & \textbf{34.08} \\
    HOIGen~\cite{guo2024HOIGen}     & \ding{51} & \textbf{31.01} & 34.57 & 33.86 \\
    \midrule
    \multicolumn{4}{@{}l@{}}{\textit{Open-vocabulary Methods}} \\
    \hline
    THID~\cite{wang2022_THID}   & \ding{55}    & 15.53 & 24.32 & 22.38 \\
    CMD-SE~\cite{lei2024CMD-SE} & \ding{55} & 16.70 & 23.95 & 22.35 \\
    \modelname (Ours)           & \ding{55}  & \textbf{17.38} & \textbf{24.74} & \textbf{23.12} \\
    \bottomrule
  \end{tabular}
  \caption{Performance comparison on HICO-DET under the simulated zero-shot setting.}
  \label{tab:hico-det}
\end{table}

\begin{table}
  \centering
  \begin{tabular}{@{}lcccc@{}}
    \toprule
    Method & Non-rare & Rare & Unseen & Full \\
    \midrule
    QPIC~\cite{tamura2021qpic} & 16.95 & 10.84 & 6.21 & 11.12  \\
    GEN-VLKT~\cite{liao2022gen} & 20.91 & 10.41 & - & 10.87 \\
    MP-HOI~\cite{yang2024MPHOI} & 20.28 & 14.78 & - & 12.61 \\
    THID~\cite{wang2022_THID}  & 17.67  & 12.82 & 10.04 & 13.26 \\
    CMD-SE~\cite{lei2024CMD-SE} &  21.46 & 14.64 & 10.70 & 15.26 \\
    \modelname (Ours) & \textbf{22.84} & \textbf{16.74}  & \textbf{11.02} & \textbf{16.74} \\    
    \bottomrule
  \end{tabular}
  \caption{Performance comparison on SWIG-HOI in terms of mAP.}
  \label{tab:swig-hoi}
\end{table}

\subsection{Ablation Study}
\label{subsec:ablation}

\noindent \textbf{Different Component.}
We empirically investigate the impact of different components of the proposed method on performance using the SWIG-HOI dataset, as shown in \cref{tab:ablation-module}. Our analysis begins with a baseline model that utilizes common prompts, built upon the model described in \cref{subsec:preliminary}. Adding INP results in substantial improvements across all dataset splits, indicating that these prompts effectively bridge the gap between image-level pre-training and the more granular region-level interaction detection. Furthermore, we observe that incorporating the CC module yields additional performance gains, particularly on the unseen split, demonstrating the enhanced generalizability of the calibrated structured semantic space.

\begin{table}
  \centering
  \begin{tabular}{@{}cccccc@{}}
    \toprule
      INP   &  CC  & Non-rare & Rare & Unseen & Full \\
    \midrule
            &      &  19.77  &  14.53 &  9.22  & 14.43 \\
     \ding{51} &   &  21.43  &  15.48 &  10.15 & 15.54 \\
      & \ding{51}  &  21.41  &  15.02 &  10.12 & 15.30 \\
    \ding{51} & \ding{51} & \textbf{22.84} & \textbf{16.74}  & \textbf{11.02} & \textbf{16.74} \\ 
    \bottomrule
  \end{tabular}
  \caption{Ablations of different modules of our \modelname on the SWIG-HOI dataset. INP: Interaction-aware Prompt. CC: Concept Calibration.}
  \vspace{-1em}
  \label{tab:ablation-module}
\end{table}

\noindent \textbf{Interaction-aware Prompts.}
In \cref{tab:ablation-interaction-prompt}, we investigate the impact of different sizes and the selection mechanism of interaction-aware prompts. As shown in the first row, when the model is only equipped with common prompts, it achieves a mAP of 15.30 on the full split. In the second row, introducing interaction-aware prompts without the selection mechanism leads to a modest improvement of 0.68 mAP. The third row demonstrates that when the model is provided with our proposed interaction-aware prompts, specifically tailored for each scene, it achieves the best performance across all splits. Finally, the last two rows show that adjusting the prompt size (either too large or too small) leads to slight performance degradation, likely due to insufficient or redundant interaction patterns.

\begin{table}
  \centering
  \begin{tabular}{@{}cccccc@{}}
    \toprule
      Size & Sel. & Non-rare & Rare & Unseen & Full \\
    \midrule
    0 &  & 21.41 & 15.02 & 10.12 & 15.30  \\ 
    128 &  & 22.13 & 15.60 & 10.41 & 15.98  \\ 
    128 & \ding{51} & \textbf{22.84} & \textbf{16.74}  & \textbf{11.02} & \textbf{16.74} \\ 
    64  & \ding{51} & 22.58 & 15.92 & 10.73 & 16.31  \\ 
    256 & \ding{51} & 22.06 & 15.90 & 10.75 & 16.18 \\ 
    \bottomrule
  \end{tabular}
  \caption{Ablations of the interaction-aware prompt. Sel.: selection mechanism.}
  \label{tab:ablation-interaction-prompt}
\end{table}

\begin{table}
  \centering
  \begin{tabular}{@{}cccccc@{}}
    \toprule
     CE  &  CT  & Non-rare & Rare & Unseen & Full \\
    \midrule
    CLIP & HOI name & 21.94 & 16.12 & 10.85 & 16.17 \\
    CLIP & desc. & 22.57  & 15.61 & 10.32 & 15.93 \\ 
    IE & desc. & \textbf{22.84} & \textbf{16.74}  & \textbf{11.02} & \textbf{16.74} \\ 
    \bottomrule
  \end{tabular}
  \caption{Ablations of the clustering design. CE: Cluster Encoder. CT: Cluster Text. IE: Instructor Embedding.}
  \label{tab:ablation-clustering}
\end{table}

\begin{table}
  \centering
  \begin{tabular}{@{}cccccc@{}}
    \toprule
    Strategy  & Non-rare & Rare & Unseen & Full \\
    \midrule 
    Easy   &   21.87 & 15.82 & 10.66 & 15.95 \\
    Random &   21.88 & 16.01 & 10.80 & 16.15 \\
    Hard   & \textbf{22.84} & \textbf{16.74}  & \textbf{11.02} & \textbf{16.74} \\ 
    \bottomrule
  \end{tabular}
  \caption{Ablations of the negative sampling strategy.}
  \vspace{-1em}
  \label{tab:negative-sampling}
\end{table}

\noindent \textbf{The Design of Calibrated Clustering.}
We explore clustering strategies for fine-grained semantic calibration in \cref{tab:ablation-clustering}. As shown in the first two lines of \cref{tab:ablation-clustering}, using CLIP embeddings of HOI descriptions results in a 0.81 mAP drop compared to using HOI names, indicating the CLIP text encoder struggles with nuanced interactions. 
Our approach, clustering visual description embeddings from the T5 space, achieves the best performance across all splits. This ablation study emphasizes the importance of visual similarity-based concept sampling for improved detection.

\noindent \textbf{Negative Sampling Strategy.}
In \cref{tab:negative-sampling}, we compare different negative sampling strategies. The ``Easy" strategy selects negative categories from clusters that do not contain any categories in the current minibatch, while the ``Random" strategy randomly selects negative categories regardless of their cluster membership.
Both ``Easy" and ``Random" show similar performance but are less effective, as they don't fully capture the complexities needed for optimal learning. The ``Hard" strategy leads to the best results by challenging the model with more difficult examples.

\begin{figure}[ht]
  \centering
  \begin{subfigure}{0.45\linewidth}
    \centering
    \includegraphics[width=\linewidth]{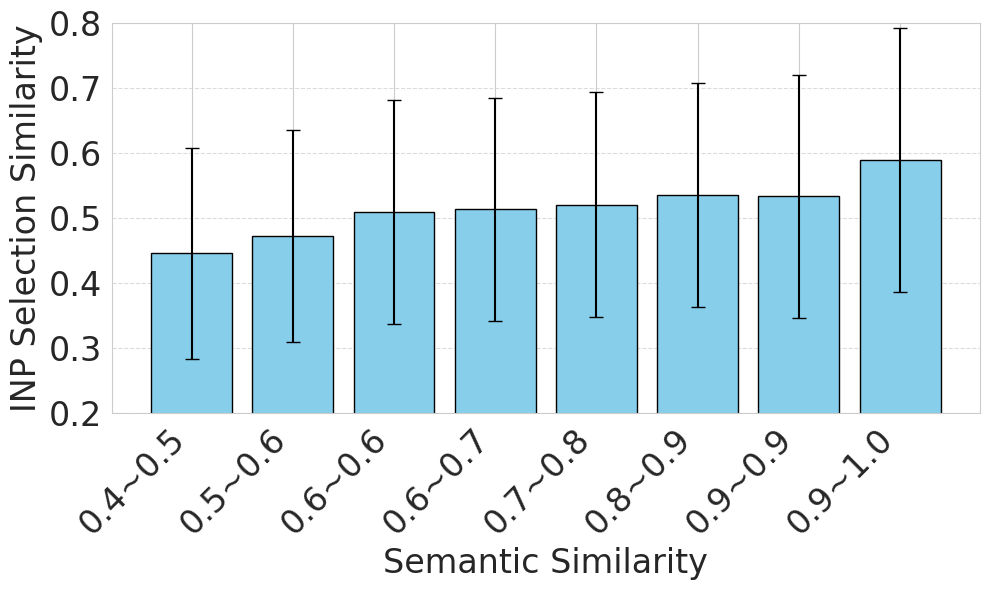}
    \caption{Inter-class semantic similarity vs INP selection similarity.}
    \label{fig:inter}
  \end{subfigure}
  \begin{subfigure}{0.45\linewidth}
    \centering
    \includegraphics[width=\linewidth]{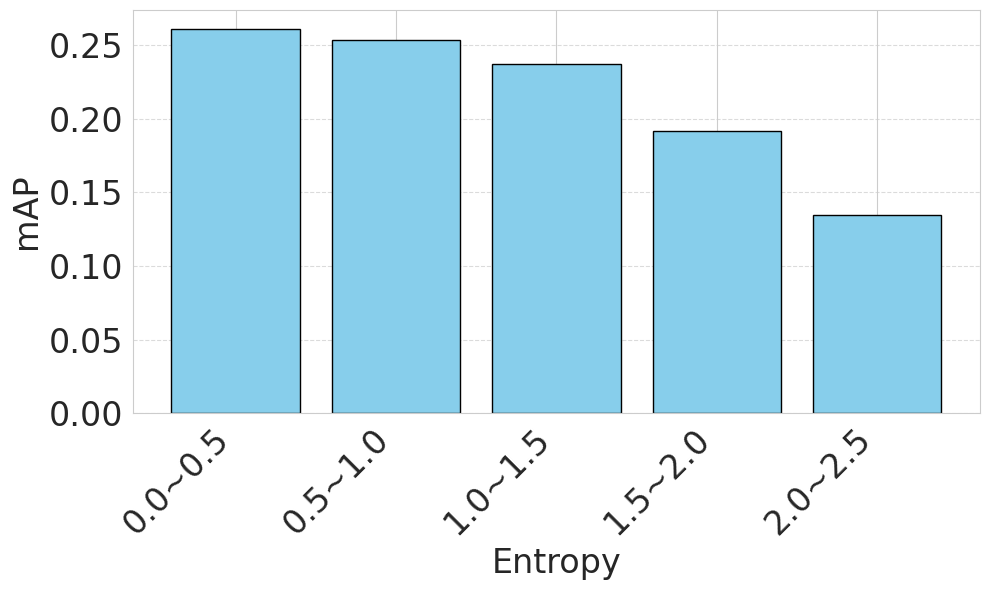}
    \caption{Intra-class INP selection entropy vs performance.}
    \label{fig:intra}
  \end{subfigure}
  \caption{Analysis on INteraction-aware Prompt~(INP).}
  \label{fig:INP_analysis}
\end{figure}

\begin{figure}[hbt]
  \centering
   \begin{subfigure}{0.49\linewidth}
        \includegraphics[width=0.49\linewidth]{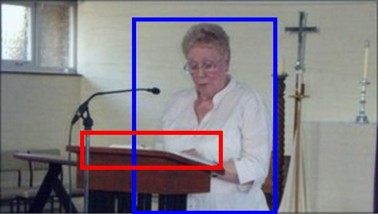}
        \includegraphics[width=0.49\linewidth]{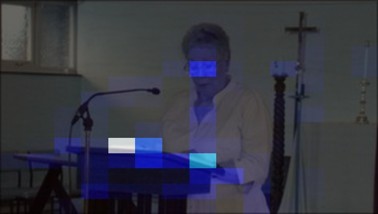}
        \caption{Reading.}
        \label{fig:read}
   \end{subfigure}
   \hfill
   \begin{subfigure}{0.49\linewidth}
        \includegraphics[width=0.49\linewidth]{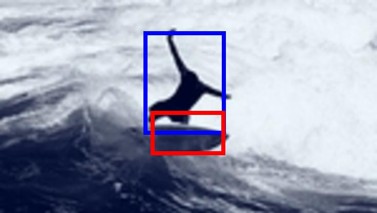}
        \includegraphics[width=0.49\linewidth]{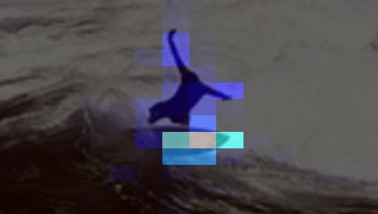}
        \caption{Surfing.}
        \label{fig:surf}
   \end{subfigure}
   \hfill
   \begin{subfigure}{0.49\linewidth}
        \includegraphics[width=0.49\linewidth]{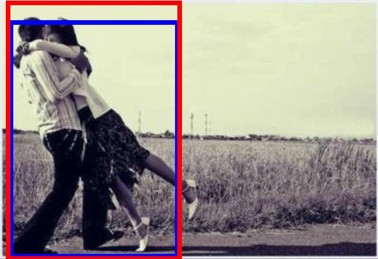}
        \includegraphics[width=0.49\linewidth]{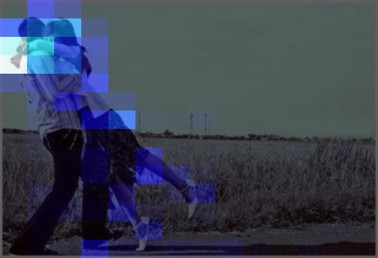}
        \caption{Hugging.}
        \label{fig:hug}
   \end{subfigure}
   \hfill
   \begin{subfigure}{0.49\linewidth}
        \includegraphics[width=0.49\linewidth]{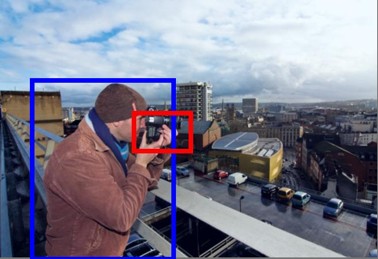}
        \includegraphics[width=0.49\linewidth]{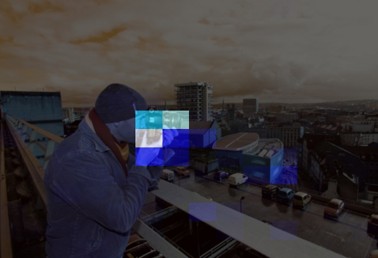}
        \caption{Photographing.}
        \label{fig:photograph}
   \end{subfigure}
   \hfill
   \caption{Qualitative examples.}
   \label{fig:vis}
\end{figure}

\noindent \textbf{Analysis of the interaction-aware prompt.}
In open-vocabulary settings, assigning unique prompts to each interaction is impractical due to the large interaction space and long-tail distribution. To overcome this, we use a compact set of INPs, where semantically similar interactions share prompts. As seen in~\cref{fig:inter}, related HOI classes often select the same INPs, indicating effective pattern capture. Lower entropy in INP selection correlates with higher mAP scores (\cref{fig:intra}), demonstrating that INPs improve the model's generalization to unseen interactions in open-world scenarios.

\subsection{Qualitative Results}
\label{subsec:qualitative}
To showcase the effectiveness of our method, we present visualizations of the prediction results in~\cref{fig:vis}. Our model demonstrates a strong ability to focus attention on key interaction regions, as evidenced by several examples. For instance, in~\cref{fig:read}, it accurately highlights the eye region during reading. Similarly, in~\cref{fig:surf}, the model emphasizes the extended arms of a person while surfing. Moreover, our model is capable of detecting interactions with relatively small objects, such as the camera in~\cref{fig:photograph} and the partially occluded book in~\cref{fig:read}.

%% file: sec/5_conclusion.tex
\section{Conclusion}
\label{sec:conclusion}
We propose interaction-adaptive prompts and concept calibration to address open-vocabulary HOI detection. Unlike prior methods hindered by the distribution gap between region-level interaction detection and image-level classification, our approach adaptively transfers interaction-aware knowledge to bridge this gap, improving model adaptability across diverse HOI scenes. We also introduce a language model-guided HOI concept calibration to capture intra-modal relationships, enhancing the model's ability to differentiate HOI concepts. A hard negative sampling strategy further enables better differentiation of visually similar yet semantically distinct actions. Experiments show significant improvements over state-of-the-art models on HICO-DET and SWIG-HOI datasets.

\noindent\textbf{Acknowledgements.}This work was supported by the
grants from the National Natural Science Foundation of
China (62372014, 62525201, 62132001, 62432001), Beijing Nova Program and Beijing Natural Science Foundation (4252040, L247006).

%% file: main.bbl
\begin{thebibliography}{105}
\providecommand{\natexlab}[1]{#1}
\providecommand{\url}[1]{\texttt{#1}}
\expandafter\ifx\csname urlstyle\endcsname\relax
  \providecommand{\doi}[1]{doi: #1}\else
  \providecommand{\doi}{doi: \begingroup \urlstyle{rm}\Url}\fi

\bibitem[Bansal et~al.(2020)Bansal, Rambhatla, Shrivastava, and Chellappa]{bansal2020func}
Ankan Bansal, Sai~Saketh Rambhatla, Abhinav Shrivastava, and Rama Chellappa.
\newblock Detecting human-object interactions via functional generalization.
\newblock In \emph{Proceedings of the AAAI Conference on Artificial Intelligence}, pages 10460--10469, 2020.

\bibitem[Caba~Heilbron et~al.(2015)Caba~Heilbron, Escorcia, Ghanem, and Carlos~Niebles]{caba2015activitynet}
Fabian Caba~Heilbron, Victor Escorcia, Bernard Ghanem, and Juan Carlos~Niebles.
\newblock Activitynet: A large-scale video benchmark for human activity understanding.
\newblock In \emph{Proceedings of the ieee conference on computer vision and pattern recognition}, pages 961--970, 2015.

\bibitem[Cao et~al.(2023{\natexlab{a}})Cao, Tang, Su, Song, You, Lu, and Xu]{cao2023UniHOI}
Yichao Cao, Qingfei Tang, Xiu Su, Chen Song, Shan You, Xiaobo Lu, and Chang Xu.
\newblock Detecting any human-object interaction relationship: Universal hoi detector with spatial prompt learning on foundation models, 2023{\natexlab{a}}.

\bibitem[Cao et~al.(2023{\natexlab{b}})Cao, Tang, Yang, Su, You, Lu, and Xu]{cao2023RmLR}
Yichao Cao, Qingfei Tang, Feng Yang, Xiu Su, Shan You, Xiaobo Lu, and Chang Xu.
\newblock Re-mine, learn and reason: Exploring the cross-modal semantic correlations for language-guided hoi detection.
\newblock In \emph{Proceedings of the IEEE/CVF International Conference on Computer Vision}, pages 23492--23503, 2023{\natexlab{b}}.

\bibitem[Carion et~al.(2020)Carion, Massa, Synnaeve, Usunier, Kirillov, and Zagoruyko]{carion2020DETR}
Nicolas Carion, Francisco Massa, Gabriel Synnaeve, Nicolas Usunier, Alexander Kirillov, and Sergey Zagoruyko.
\newblock End-to-end object detection with transformers.
\newblock In \emph{European conference on computer vision}, pages 213--229. Springer, 2020.

\bibitem[Chao et~al.(2018)Chao, Liu, Liu, Zeng, and Deng]{chao2018HICO-DET}
Yu-Wei Chao, Yunfan Liu, Xieyang Liu, Huayi Zeng, and Jia Deng.
\newblock Learning to detect human-object interactions.
\newblock In \emph{2018 ieee winter conference on applications of computer vision (wacv)}, pages 381--389. IEEE, 2018.

\bibitem[Chen and Yanai(2023)]{chen2023qahoi}
Junwen Chen and Keiji Yanai.
\newblock Qahoi: Query-based anchors for human-object interaction detection.
\newblock In \emph{2023 18th International Conference on Machine Vision and Applications (MVA)}, pages 1--5. IEEE, 2023.

\bibitem[Chen et~al.(2021)Chen, Liao, Liu, Chen, Wang, and Qian]{chen2021reformulating}
Mingfei Chen, Yue Liao, Si Liu, Zhiyuan Chen, Fei Wang, and Chen Qian.
\newblock Reformulating hoi detection as adaptive set prediction.
\newblock In \emph{Proceedings of the IEEE/CVF Conference on Computer Vision and Pattern Recognition}, pages 9004--9013, 2021.

\bibitem[Cheng et~al.(2024)Cheng, Duan, Wang, and Chen]{cheng2024parallel}
Yamin Cheng, Hancong Duan, Chen Wang, and Zhijun Chen.
\newblock Parallel disentangling network for human--object interaction detection.
\newblock \emph{Pattern Recognition}, 146:\penalty0 110021, 2024.

\bibitem[Dong et~al.(2022)Dong, Li, Xu, Zhang, Yan, Zhong, and Zou]{dong2022category}
Leizhen Dong, Zhimin Li, Kunlun Xu, Zhijun Zhang, Luxin Yan, Sheng Zhong, and Xu Zou.
\newblock Category-aware transformer network for better human-object interaction detection.
\newblock In \emph{Proceedings of the IEEE/CVF Conference on Computer Vision and Pattern Recognition}, pages 19538--19547, 2022.

\bibitem[Du et~al.(2025)Du, Wang, Sun, Wang, Liao, Zhang, Ding, Wang, Wang, and Liu]{du2025lami}
Penghui Du, Yu Wang, Yifan Sun, Luting Wang, Yue Liao, Gang Zhang, Errui Ding, Yan Wang, Jingdong Wang, and Si Liu.
\newblock Lami-detr: Open-vocabulary detection with language model instruction.
\newblock In \emph{European Conference on Computer Vision}, pages 312--328. Springer, 2025.

\bibitem[Fang et~al.(2021)Fang, Xie, Shao, and Lu]{fang2020dirv}
Hao-Shu Fang, Yichen Xie, Dian Shao, and Cewu Lu.
\newblock Dirv: Dense interaction region voting for end-to-end human-object interaction detection.
\newblock In \emph{The AAAI Conference on Artificial Intelligence (AAAI)}, 2021.

\bibitem[Feichtenhofer et~al.(2017)Feichtenhofer, Pinz, and Wildes]{feichtenhofer2017spatiotemporal}
Christoph Feichtenhofer, Axel Pinz, and Richard~P Wildes.
\newblock Spatiotemporal multiplier networks for video action recognition.
\newblock In \emph{Proceedings of the IEEE conference on computer vision and pattern recognition}, pages 4768--4777, 2017.

\bibitem[Gao et~al.(2018)Gao, Zou, and Huang]{gao2018ican}
Chen Gao, Yuliang Zou, and Jia-Bin Huang.
\newblock ican: Instance-centric attention network for human-object interaction detection.
\newblock \emph{arXiv preprint arXiv:1808.10437}, 2018.

\bibitem[Gao et~al.(2020)Gao, Xu, Zou, and Huang]{gao2020drg}
Chen Gao, Jiarui Xu, Yuliang Zou, and Jia-Bin Huang.
\newblock Drg: Dual relation graph for human-object interaction detection.
\newblock In \emph{European Conference on Computer Vision}, pages 696--712. Springer, 2020.

\bibitem[Gao et~al.(2025)Gao, Yin, Hua, Peng, Liang, Ma, Guo, and Liu]{gao2025conmo}
Jiayi Gao, Zijin Yin, Changcheng Hua, Yuxin Peng, Kongming Liang, Zhanyu Ma, Jun Guo, and Yang Liu.
\newblock Conmo: Controllable motion disentanglement and recomposition for zero-shot motion transfer.
\newblock In \emph{Proceedings of the Computer Vision and Pattern Recognition Conference}, pages 7191--7200, 2025.

\bibitem[Gkioxari et~al.(2018)Gkioxari, Girshick, Doll{\'a}r, and He]{gkioxari2018detecting}
Georgia Gkioxari, Ross Girshick, Piotr Doll{\'a}r, and Kaiming He.
\newblock Detecting and recognizing human-object interactions.
\newblock In \emph{Proceedings of the IEEE conference on computer vision and pattern recognition}, pages 8359--8367, 2018.

\bibitem[Guo et~al.(2024)Guo, Liu, Li, Wang, and Jia]{guo2024HOIGen}
Yixin Guo, Yu Liu, Jianghao Li, Weimin Wang, and Qi Jia.
\newblock Unseen no more: Unlocking the potential of clip for generative zero-shot hoi detection.
\newblock In \emph{Proceedings of the 32nd ACM International Conference on Multimedia}, pages 1711--1720, 2024.

\bibitem[Gupta et~al.(2019)Gupta, Schwing, and Hoiem]{gupta2019no}
Tanmay Gupta, Alexander Schwing, and Derek Hoiem.
\newblock No-frills human-object interaction detection: Factorization, layout encodings, and training techniques.
\newblock In \emph{Proceedings of the IEEE/CVF International Conference on Computer Vision}, pages 9677--9685, 2019.

\bibitem[Hou et~al.(2020)Hou, Peng, Qiao, and Tao]{hou2020VCL}
Zhi Hou, Xiaojiang Peng, Yu Qiao, and Dacheng Tao.
\newblock Visual compositional learning for human-object interaction detection.
\newblock In \emph{European Conference on Computer Vision}, pages 584--600. Springer, 2020.

\bibitem[Hou et~al.(2021{\natexlab{a}})Hou, Yu, Qiao, Peng, and Tao]{hou2021ATL}
Zhi Hou, Baosheng Yu, Yu Qiao, Xiaojiang Peng, and Dacheng Tao.
\newblock Affordance transfer learning for human-object interaction detection.
\newblock In \emph{Proceedings of the IEEE/CVF Conference on Computer Vision and Pattern Recognition}, pages 495--504, 2021{\natexlab{a}}.

\bibitem[Hou et~al.(2021{\natexlab{b}})Hou, Yu, Qiao, Peng, and Tao]{hou2021FCL}
Zhi Hou, Baosheng Yu, Yu Qiao, Xiaojiang Peng, and Dacheng Tao.
\newblock Detecting human-object interaction via fabricated compositional learning.
\newblock In \emph{Proceedings of the IEEE/CVF Conference on Computer Vision and Pattern Recognition}, pages 14646--14655, 2021{\natexlab{b}}.

\bibitem[Iftekhar et~al.(2022{\natexlab{a}})Iftekhar, Chen, Kundu, Li, Tighe, and Modolo]{iftekhar2022SSRT}
ASM Iftekhar, Hao Chen, Kaustav Kundu, Xinyu Li, Joseph Tighe, and Davide Modolo.
\newblock What to look at and where: Semantic and spatial refined transformer for detecting human-object interactions.
\newblock In \emph{Proceedings of the IEEE/CVF Conference on Computer Vision and Pattern Recognition}, pages 5353--5363, 2022{\natexlab{a}}.

\bibitem[Iftekhar et~al.(2022{\natexlab{b}})Iftekhar, Chen, Kundu, Li, Tighe, and Modolo]{iftekhar2022look}
ASM Iftekhar, Hao Chen, Kaustav Kundu, Xinyu Li, Joseph Tighe, and Davide Modolo.
\newblock What to look at and where: Semantic and spatial refined transformer for detecting human-object interactions.
\newblock In \emph{Proceedings of the IEEE/CVF Conference on Computer Vision and Pattern Recognition}, pages 5353--5363, 2022{\natexlab{b}}.

\bibitem[Jiang et~al.(2024)Jiang, Ren, Tian, Qu, Wang, and Liu]{jiang2024SCTC}
Weibo Jiang, Weihong Ren, Jiandong Tian, Liangqiong Qu, Zhiyong Wang, and Honghai Liu.
\newblock Exploring self-and cross-triplet correlations for human-object interaction detection.
\newblock In \emph{Proceedings of the AAAI Conference on Artificial Intelligence}, pages 2543--2551, 2024.

\bibitem[Jin et~al.(2024)Jin, Jiang, Huang, Lu, and Lu]{jin2024llms}
Sheng Jin, Xueying Jiang, Jiaxing Huang, Lewei Lu, and Shijian Lu.
\newblock Llms meet vlms: Boost open vocabulary object detection with fine-grained descriptors, 2024.

\bibitem[Kato et~al.(2018)Kato, Li, and Gupta]{kato2018compositional}
Keizo Kato, Yin Li, and Abhinav Gupta.
\newblock Compositional learning for human object interaction.
\newblock In \emph{Proceedings of the European Conference on Computer Vision (ECCV)}, pages 234--251, 2018.

\bibitem[Kaul et~al.(2023)Kaul, Xie, and Zisserman]{kaul2023multimodal}
Prannay Kaul, Weidi Xie, and Andrew Zisserman.
\newblock Multi-modal classifiers for open-vocabulary object detection, 2023.

\bibitem[Kim et~al.(2020)Kim, Choi, Kang, and Kim]{kim2020uniondet}
Bumsoo Kim, Taeho Choi, Jaewoo Kang, and Hyunwoo~J Kim.
\newblock Uniondet: Union-level detector towards real-time human-object interaction detection.
\newblock In \emph{European Conference on Computer Vision}, pages 498--514. Springer, 2020.

\bibitem[Kim et~al.(2021)Kim, Lee, Kang, Kim, and Kim]{kim2021hotr}
Bumsoo Kim, Junhyun Lee, Jaewoo Kang, Eun-Sol Kim, and Hyunwoo~J Kim.
\newblock Hotr: End-to-end human-object interaction detection with transformers.
\newblock In \emph{Proceedings of the IEEE/CVF Conference on Computer Vision and Pattern Recognition}, pages 74--83, 2021.

\bibitem[Kim et~al.(2022)Kim, Mun, On, Shin, Lee, and Kim]{kim2022mstr}
Bumsoo Kim, Jonghwan Mun, Kyoung-Woon On, Minchul Shin, Junhyun Lee, and Eun-Sol Kim.
\newblock Mstr: Multi-scale transformer for end-to-end human-object interaction detection.
\newblock In \emph{Proceedings of the IEEE/CVF Conference on Computer Vision and Pattern Recognition}, pages 19578--19587, 2022.

\bibitem[Kim et~al.(2023{\natexlab{a}})Kim, Angelova, and Kuo]{kim2023region}
Dahun Kim, Anelia Angelova, and Weicheng Kuo.
\newblock Region-aware pretraining for open-vocabulary object detection with vision transformers.
\newblock In \emph{Proceedings of the IEEE/CVF conference on computer vision and pattern recognition}, pages 11144--11154, 2023{\natexlab{a}}.

\bibitem[Kim et~al.(2023{\natexlab{b}})Kim, Jung, and Cho]{Kim_2023_CVPR}
Sanghyun Kim, Deunsol Jung, and Minsu Cho.
\newblock Relational context learning for human-object interaction detection.
\newblock In \emph{Proceedings of the IEEE/CVF Conference on Computer Vision and Pattern Recognition (CVPR)}, pages 2925--2934, 2023{\natexlab{b}}.

\bibitem[Lei et~al.(2023)Lei, Caba, Chen, Ji, Peng, and Liu]{ting2023hoi}
Ting Lei, Fabian Caba, Qingchao Chen, Hailin Ji, Yuxin Peng, and Yang Liu.
\newblock Efficient adaptive human-object interaction detection with concept-guided memory.
\newblock In \emph{ICCV}. IEEE, 2023.

\bibitem[Lei et~al.(2024{\natexlab{a}})Lei, Yin, and Liu]{lei2024CMD-SE}
Ting Lei, Shaofeng Yin, and Yang Liu.
\newblock Exploring the potential of large foundation models for open-vocabulary hoi detection.
\newblock In \emph{Proceedings of the IEEE/CVF Conference on Computer Vision and Pattern Recognition}, pages 16657--16667, 2024{\natexlab{a}}.

\bibitem[Lei et~al.(2024{\natexlab{b}})Lei, Yin, Peng, and Liu]{Lei_2024_eccv}
Ting Lei, Shaofeng Yin, Yuxin Peng, and Yang Liu.
\newblock Exploring conditional multi-modal prompts for zero-shot hoi detection.
\newblock In \emph{European Conference on Computer Vision}, pages 1--19. Springer, 2024{\natexlab{b}}.

\bibitem[Li et~al.(2018)Li, Farkhoor, Liu, and Yosinski]{li2018measuring}
Chunyuan Li, Heerad Farkhoor, Rosanne Liu, and Jason Yosinski.
\newblock Measuring the intrinsic dimension of objective landscapes.
\newblock \emph{arXiv preprint arXiv:1804.08838}, 2018.

\bibitem[Li et~al.(2023{\natexlab{a}})Li, Lai, Gao, Ma, Quan, and Chen]{li2023sqab}
Junkai Li, Huicheng Lai, Guxue Gao, Jun Ma, Hutuo Quan, and Dongji Chen.
\newblock Sqab: Specific query anchor boxes for human--object interaction detection.
\newblock \emph{Displays}, 80:\penalty0 102570, 2023{\natexlab{a}}.

\bibitem[Li et~al.(2023{\natexlab{b}})Li, Xiao, Chen, Shao, Zhuang, and Chen]{li2023zeroshot}
Lin Li, Jun Xiao, Guikun Chen, Jian Shao, Yueting Zhuang, and Long Chen.
\newblock Zero-shot visual relation detection via composite visual cues from large language models, 2023{\natexlab{b}}.

\bibitem[Li et~al.(2024{\natexlab{a}})Li, Wang, and Yang]{li2024diffusionHOI}
Liulei Li, Wenguan Wang, and Yi Yang.
\newblock Human-object interaction detection collaborated with large relation-driven diffusion models.
\newblock \emph{arXiv preprint arXiv:2410.20155}, 2024{\natexlab{a}}.

\bibitem[Li et~al.(2024{\natexlab{b}})Li, Wei, Wang, and Yang]{li2024logichoi}
Liulei Li, Jianan Wei, Wenguan Wang, and Yi Yang.
\newblock Neural-logic human-object interaction detection.
\newblock \emph{Advances in Neural Information Processing Systems}, 36, 2024{\natexlab{b}}.

\bibitem[Li et~al.(2024{\natexlab{c}})Li, Zhang, Lin, Chen, and He]{li2024pixels}
Rongjie Li, Songyang Zhang, Dahua Lin, Kai Chen, and Xuming He.
\newblock From pixels to graphs: Open-vocabulary scene graph generation with vision-language models, 2024{\natexlab{c}}.

\bibitem[Li et~al.(2019)Li, Zhou, Huang, Xu, Ma, Fang, Wang, and Lu]{li2019transferable}
Yong-Lu Li, Siyuan Zhou, Xijie Huang, Liang Xu, Ze Ma, Hao-Shu Fang, Yanfeng Wang, and Cewu Lu.
\newblock Transferable interactiveness knowledge for human-object interaction detection.
\newblock In \emph{Proceedings of the IEEE/CVF Conference on Computer Vision and Pattern Recognition}, pages 3585--3594, 2019.

\bibitem[Li et~al.(2020)Li, Xu, Liu, Huang, Xu, Wang, Fang, Ma, Chen, and Lu]{li2020pastanet}
Yong-Lu Li, Liang Xu, Xinpeng Liu, Xijie Huang, Yue Xu, Shiyi Wang, Hao-Shu Fang, Ze Ma, Mingyang Chen, and Cewu Lu.
\newblock Pastanet: Toward human activity knowledge engine.
\newblock In \emph{Proceedings of the IEEE/CVF Conference on Computer Vision and Pattern Recognition}, pages 382--391, 2020.

\bibitem[Li et~al.(2022)Li, Zou, Zhao, Li, and Zhong]{li2022improving}
Zhimin Li, Cheng Zou, Yu Zhao, Boxun Li, and Sheng Zhong.
\newblock Improving human-object interaction detection via phrase learning and label composition.
\newblock In \emph{Proceedings of the AAAI Conference on Artificial Intelligence}, pages 1509--1517, 2022.

\bibitem[Li et~al.(2024{\natexlab{d}})Li, Li, Ding, and Xu]{li2024disentangledPretrain}
Zhuolong Li, Xingao Li, Changxing Ding, and Xiangmin Xu.
\newblock Disentangled pre-training for human-object interaction detection.
\newblock In \emph{Proceedings of the IEEE/CVF Conference on Computer Vision and Pattern Recognition}, pages 28191--28201, 2024{\natexlab{d}}.

\bibitem[Liao et~al.(2020)Liao, Liu, Wang, Chen, Qian, and Feng]{liao2020ppdm}
Yue Liao, Si Liu, Fei Wang, Yanjie Chen, Chen Qian, and Jiashi Feng.
\newblock Ppdm: Parallel point detection and matching for real-time human-object interaction detection.
\newblock In \emph{Proceedings of the IEEE/CVF Conference on Computer Vision and Pattern Recognition}, pages 482--490, 2020.

\bibitem[Liao et~al.(2022)Liao, Zhang, Lu, Wang, Li, and Liu]{liao2022gen}
Yue Liao, Aixi Zhang, Miao Lu, Yongliang Wang, Xiaobo Li, and Si Liu.
\newblock Gen-vlkt: Simplify association and enhance interaction understanding for hoi detection.
\newblock In \emph{Proceedings of the IEEE/CVF Conference on Computer Vision and Pattern Recognition}, pages 20123--20132, 2022.

\bibitem[Lin et~al.(2014)Lin, Maire, Belongie, Hays, Perona, Ramanan, Doll{\'a}r, and Zitnick]{lin2014COCO}
Tsung-Yi Lin, Michael Maire, Serge Belongie, James Hays, Pietro Perona, Deva Ramanan, Piotr Doll{\'a}r, and C~Lawrence Zitnick.
\newblock Microsoft coco: Common objects in context.
\newblock In \emph{Computer Vision--ECCV 2014: 13th European Conference, Zurich, Switzerland, September 6-12, 2014, Proceedings, Part V 13}, pages 740--755. Springer, 2014.

\bibitem[Liu et~al.(2022)Liu, Li, Wu, Tai, Lu, and Tang]{liu2022interactiveness_field}
Xinpeng Liu, Yong-Lu Li, Xiaoqian Wu, Yu-Wing Tai, Cewu Lu, and Chi-Keung Tang.
\newblock Interactiveness field in human-object interactions.
\newblock In \emph{Proceedings of the IEEE/CVF Conference on Computer Vision and Pattern Recognition}, pages 20113--20122, 2022.

\bibitem[Luo et~al.(2024)Luo, Ren, Jiang, Chen, Wang, Han, and Liu]{luo2024SIC}
Jinguo Luo, Weihong Ren, Weibo Jiang, Xi'ai Chen, Qiang Wang, Zhi Han, and Honghai Liu.
\newblock Discovering syntactic interaction clues for human-object interaction detection.
\newblock In \emph{Proceedings of the IEEE/CVF Conference on Computer Vision and Pattern Recognition}, pages 28212--28222, 2024.

\bibitem[Mao et~al.(2023)Mao, Deng, Zhou, Li, Fang, and Li]{mao2023clip4hoi}
Yunyao Mao, Jiajun Deng, Wengang Zhou, Li Li, Yao Fang, and Houqiang Li.
\newblock Clip4hoi: towards adapting clip for practical zero-shot hoi detection.
\newblock \emph{Advances in Neural Information Processing Systems}, 36:\penalty0 45895--45906, 2023.

\bibitem[Menon and Vondrick(2022)]{menon2022visual}
Sachit Menon and Carl Vondrick.
\newblock Visual classification via description from large language models.
\newblock \emph{arXiv preprint arXiv:2210.07183}, 2022.

\bibitem[Mo and Liu(2024)]{mo2024bridging}
Wentao Mo and Yang Liu.
\newblock Bridging the gap between 2d and 3d visual question answering: A fusion approach for 3d vqa.
\newblock In \emph{Proceedings of the AAAI Conference on Artificial Intelligence}, pages 4261--4268, 2024.

\bibitem[Ni et~al.(2021)Ni, Abrego, Constant, Ma, Hall, Cer, and Yang]{ni2021sentence}
Jianmo Ni, Gustavo~Hernandez Abrego, Noah Constant, Ji Ma, Keith~B Hall, Daniel Cer, and Yinfei Yang.
\newblock Sentence-t5: Scalable sentence encoders from pre-trained text-to-text models.
\newblock \emph{arXiv preprint arXiv:2108.08877}, 2021.

\bibitem[Ning et~al.(2023)Ning, Qiu, Liu, and He]{ning2023hoiclip}
Shan Ning, Longtian Qiu, Yongfei Liu, and Xuming He.
\newblock Hoiclip: Efficient knowledge transfer for hoi detection with vision-language models.
\newblock In \emph{Proceedings of the IEEE/CVF Conference on Computer Vision and Pattern Recognition}, pages 23507--23517, 2023.

\bibitem[Park et~al.(2023)Park, Park, and Lee]{Park_2023_CVPR}
Jeeseung Park, Jin-Woo Park, and Jong-Seok Lee.
\newblock Viplo: Vision transformer based pose-conditioned self-loop graph for human-object interaction detection.
\newblock In \emph{Proceedings of the IEEE/CVF Conference on Computer Vision and Pattern Recognition (CVPR)}, pages 17152--17162, 2023.

\bibitem[Pratt et~al.(2023)Pratt, Covert, Liu, and Farhadi]{pratt2023does}
Sarah Pratt, Ian Covert, Rosanne Liu, and Ali Farhadi.
\newblock What does a platypus look like? generating customized prompts for zero-shot image classification.
\newblock In \emph{Proceedings of the IEEE/CVF International Conference on Computer Vision}, pages 15691--15701, 2023.

\bibitem[Radford et~al.(2021)Radford, Kim, Hallacy, Ramesh, Goh, Agarwal, Sastry, Askell, Mishkin, Clark, et~al.]{CLIP}
Alec Radford, Jong~Wook Kim, Chris Hallacy, Aditya Ramesh, Gabriel Goh, Sandhini Agarwal, Girish Sastry, Amanda Askell, Pamela Mishkin, Jack Clark, et~al.
\newblock Learning transferable visual models from natural language supervision.
\newblock In \emph{International Conference on Machine Learning}, pages 8748--8763. PMLR, 2021.

\bibitem[Ren et~al.(2016)Ren, He, Girshick, and Sun]{ren2016faster_rcnn}
Shaoqing Ren, Kaiming He, Ross Girshick, and Jian Sun.
\newblock Faster r-cnn: Towards real-time object detection with region proposal networks.
\newblock \emph{IEEE transactions on pattern analysis and machine intelligence}, 39\penalty0 (6):\penalty0 1137--1149, 2016.

\bibitem[Su et~al.(2022)Su, Shi, Kasai, Wang, Hu, Ostendorf, Yih, Smith, Zettlemoyer, and Yu]{INSTRUCTOR}
Hongjin Su, Weijia Shi, Jungo Kasai, Yizhong Wang, Yushi Hu, Mari Ostendorf, Wen-tau Yih, Noah~A. Smith, Luke Zettlemoyer, and Tao Yu.
\newblock One embedder, any task: Instruction-finetuned text embeddings.
\newblock 2022.

\bibitem[Tamura et~al.(2021)Tamura, Ohashi, and Yoshinaga]{tamura2021qpic}
Masato Tamura, Hiroki Ohashi, and Tomoaki Yoshinaga.
\newblock Qpic: Query-based pairwise human-object interaction detection with image-wide contextual information.
\newblock In \emph{Proceedings of the IEEE/CVF Conference on Computer Vision and Pattern Recognition}, pages 10410--10419, 2021.

\bibitem[Tian et~al.(2023)Tian, Fu, and Zhang]{tian2023transformer}
Ye Tian, Ying Fu, and Jun Zhang.
\newblock Transformer-based under-sampled single-pixel imaging.
\newblock \emph{Chinese Journal of Electronics}, 32\penalty0 (5):\penalty0 1151--1159, 2023.

\bibitem[Tu et~al.(2023)Tu, Sun, Zhai, and Shen]{Tu_2023_ICCV}
Danyang Tu, Wei Sun, Guangtao Zhai, and Wei Shen.
\newblock Agglomerative transformer for human-object interaction detection.
\newblock In \emph{Proceedings of the IEEE/CVF International Conference on Computer Vision (ICCV)}, pages 21614--21624, 2023.

\bibitem[Ulutan et~al.(2020)Ulutan, Iftekhar, and Manjunath]{ulutan2020vsgnet}
Oytun Ulutan, ASM Iftekhar, and Bangalore~S Manjunath.
\newblock Vsgnet: Spatial attention network for detecting human object interactions using graph convolutions.
\newblock In \emph{Proceedings of the IEEE/CVF conference on computer vision and pattern recognition}, pages 13617--13626, 2020.

\bibitem[Unal and Kovashka(2023)]{unal2023weaklysupervised}
Mesut~Erhan Unal and Adriana Kovashka.
\newblock Weakly-supervised hoi detection from interaction labels only and language/vision-language priors, 2023.

\bibitem[Wan and Tuytelaars(2024)]{wan2024exploiting}
Bo Wan and Tinne Tuytelaars.
\newblock Exploiting clip for zero-shot hoi detection requires knowledge distillation at multiple levels.
\newblock In \emph{Proceedings of the IEEE/CVF Winter Conference on Applications of Computer Vision}, pages 1805--1815, 2024.

\bibitem[Wan et~al.(2023)Wan, Liu, Zhou, Tuytelaars, and He]{wan2023weaklyHOI}
Bo Wan, Yongfei Liu, Desen Zhou, Tinne Tuytelaars, and Xuming He.
\newblock Weakly-supervised hoi detection via prior-guided bi-level representation learning.
\newblock \emph{arXiv preprint arXiv:2303.01313}, 2023.

\bibitem[Wang et~al.(2022{\natexlab{a}})Wang, Guo, Wong, and Kankanhalli]{wang2022chairs}
Guangzhi Wang, Yangyang Guo, Yongkang Wong, and Mohan Kankanhalli.
\newblock Chairs can be stood on: Overcoming object bias in human-object interaction detection.
\newblock In \emph{European Conference on Computer Vision}, pages 654--672. Springer, 2022{\natexlab{a}}.

\bibitem[Wang et~al.(2022{\natexlab{b}})Wang, Guo, Wong, and Kankanhalli]{wang2022distance}
Guangzhi Wang, Yangyang Guo, Yongkang Wong, and Mohan Kankanhalli.
\newblock Distance matters in human-object interaction detection.
\newblock In \emph{Proceedings of the 30th ACM International Conference on Multimedia}, pages 4546--4554, 2022{\natexlab{b}}.

\bibitem[Wang et~al.(2024{\natexlab{a}})Wang, Guo, Xu, and Kankanhalli]{wang2024bilateral}
Guangzhi Wang, Yangyang Guo, Ziwei Xu, and Mohan Kankanhalli.
\newblock Bilateral adaptation for human-object interaction detection with occlusion-robustness.
\newblock In \emph{Proceedings of the IEEE/CVF Conference on Computer Vision and Pattern Recognition}, pages 27970--27980, 2024{\natexlab{a}}.

\bibitem[Wang et~al.(2024{\natexlab{b}})Wang, Li, Chen, and Liu]{wang2024oed}
Guan Wang, Zhimin Li, Qingchao Chen, and Yang Liu.
\newblock Oed: towards one-stage end-to-end dynamic scene graph generation.
\newblock In \emph{Proceedings of the IEEE/CVF Conference on Computer Vision and Pattern Recognition}, pages 27938--27947, 2024{\natexlab{b}}.

\bibitem[Wang et~al.(2021)Wang, Yap, Ding, Wu, Yuan, and Tan]{wang2021SWIG-HOI}
Suchen Wang, Kim-Hui Yap, Henghui Ding, Jiyan Wu, Junsong Yuan, and Yap-Peng Tan.
\newblock Discovering human interactions with large-vocabulary objects via query and multi-scale detection.
\newblock In \emph{Proceedings of the IEEE/CVF International Conference on Computer Vision}, pages 13475--13484, 2021.

\bibitem[Wang et~al.(2022{\natexlab{c}})Wang, Duan, Ding, Tan, Yap, and Yuan]{wang2022_THID}
Suchen Wang, Yueqi Duan, Henghui Ding, Yap-Peng Tan, Kim-Hui Yap, and Junsong Yuan.
\newblock Learning transferable human-object interaction detectors with natural language supervision.
\newblock In \emph{CVPR}, 2022{\natexlab{c}}.

\bibitem[Wang et~al.(2024{\natexlab{c}})Wang, Teng, and Wang]{wang2024cyclehoi}
Yisen Wang, Yao Teng, and Limin Wang.
\newblock Cyclehoi: Improving human-object interaction detection with cycle consistency of detection and generation.
\newblock \emph{arXiv preprint arXiv:2407.11433}, 2024{\natexlab{c}}.

\bibitem[Wu et~al.(2024)Wu, Li, Wang, and Wang]{wu2024pose_aware}
Eastman~ZY Wu, Yali Li, Yuan Wang, and Shengjin Wang.
\newblock Exploring pose-aware human-object interaction via hybrid learning.
\newblock In \emph{Proceedings of the IEEE/CVF Conference on Computer Vision and Pattern Recognition}, pages 17815--17825, 2024.

\bibitem[Wu et~al.(2022)Wu, Gu, Shen, Lin, Chen, Sun, and Ji]{wu2022EoID}
Mingrui Wu, Jiaxin Gu, Yunhang Shen, Mingbao Lin, Chao Chen, Xiaoshuai Sun, and Rongrong Ji.
\newblock End-to-end zero-shot hoi detection via vision and language knowledge distillation.
\newblock \emph{arXiv preprint arXiv:2204.03541}, 2022.

\bibitem[Xie et~al.(2023)Xie, Zeng, Hu, Liang, and Wei]{Xie_2023_CVPR}
Chi Xie, Fangao Zeng, Yue Hu, Shuang Liang, and Yichen Wei.
\newblock Category query learning for human-object interaction classification.
\newblock In \emph{Proceedings of the IEEE/CVF Conference on Computer Vision and Pattern Recognition (CVPR)}, pages 15275--15284, 2023.

\bibitem[Xu et~al.(2019)Xu, Wong, Li, Zhao, and Kankanhalli]{xu2019learning}
Bingjie Xu, Yongkang Wong, Junnan Li, Qi Zhao, and Mohan~S Kankanhalli.
\newblock Learning to detect human-object interactions with knowledge.
\newblock In \emph{Proceedings of the IEEE/CVF Conference on Computer Vision and Pattern Recognition}, 2019.

\bibitem[Xu et~al.(2024)Xu, Chen, Peng, and Liu]{xu2024semantic}
Zhu Xu, Qingchao Chen, Yuxin Peng, and Yang Liu.
\newblock Semantic-aware human object interaction image generation.
\newblock In \emph{Forty-first International Conference on Machine Learning}, 2024.

\bibitem[Xue et~al.(2024)Xue, Liu, Xiong, Wang, Wei, Xing, and Xu]{xue2024KI2HOI}
Weiying Xue, Qi Liu, Qiwei Xiong, Yuxiao Wang, Zhenao Wei, Xiaofen Xing, and Xiangmin Xu.
\newblock Towards zero-shot human-object interaction detection via vision-language integration.
\newblock \emph{arXiv preprint arXiv:2403.07246}, 2024.

\bibitem[Yamada et~al.(2022)Yamada, Tang, and Yildirim]{yamada2022lemons}
Yutaro Yamada, Yingtian Tang, and Ilker Yildirim.
\newblock When are lemons purple? the concept association bias of clip.
\newblock \emph{arXiv preprint arXiv:2212.12043}, 2022.

\bibitem[Yang and Liu(2024)]{yang2024active}
Dejie Yang and Yang Liu.
\newblock Active object detection with knowledge aggregation and distillation from large models.
\newblock In \emph{Proceedings of the IEEE/CVF Conference on Computer Vision and Pattern Recognition}, pages 16624--16633, 2024.

\bibitem[Yang and Zou(2020)]{yang2020graph}
Dongming Yang and Yuexian Zou.
\newblock A graph-based interactive reasoning for human-object interaction detection.
\newblock \emph{arXiv preprint arXiv:2007.06925}, 2020.

\bibitem[Yang et~al.(2024{\natexlab{a}})Yang, Xu, Mo, Chen, Huang, and Liu]{yang20243d}
Dejie Yang, Zhu Xu, Wentao Mo, Qingchao Chen, Siyuan Huang, and Yang Liu.
\newblock 3d vision and language pretraining with large-scale synthetic data.
\newblock \emph{arXiv preprint arXiv:2407.06084}, 2024{\natexlab{a}}.

\bibitem[Yang et~al.(2025)Yang, Zhao, and Liu]{yang2025planllm}
Dejie Yang, Zijing Zhao, and Yang Liu.
\newblock Planllm: Video procedure planning with refinable large language models.
\newblock In \emph{Proceedings of the AAAI Conference on Artificial Intelligence}, pages 9166--9174, 2025.

\bibitem[Yang et~al.(2024{\natexlab{b}})Yang, Li, Zeng, Zhang, and Zhang]{yang2024MPHOI}
Jie Yang, Bingliang Li, Ailing Zeng, Lei Zhang, and Ruimao Zhang.
\newblock Open-world human-object interaction detection via multi-modal prompts.
\newblock In \emph{Proceedings of the IEEE/CVF Conference on Computer Vision and Pattern Recognition}, pages 16954--16964, 2024{\natexlab{b}}.

\bibitem[Yang et~al.(2023)Yang, Panagopoulou, Zhou, Jin, Callison-Burch, and Yatskar]{yang2023language}
Yue Yang, Artemis Panagopoulou, Shenghao Zhou, Daniel Jin, Chris Callison-Burch, and Mark Yatskar.
\newblock Language in a bottle: Language model guided concept bottlenecks for interpretable image classification.
\newblock In \emph{Proceedings of the IEEE/CVF Conference on Computer Vision and Pattern Recognition}, pages 19187--19197, 2023.

\bibitem[Yang et~al.(2024{\natexlab{c}})Yang, Liu, Ouyang, Duan, Zhang, He, and Li]{yang2024CaCLIP}
Zhenhao Yang, Xin Liu, Deqiang Ouyang, Guiduo Duan, Dongyang Zhang, Tao He, and Yuan-Fang Li.
\newblock Towards open-vocabulary hoi detection with calibrated vision-language models and locality-aware queries.
\newblock In \emph{Proceedings of the 32nd ACM International Conference on Multimedia}, pages 1495--1504, 2024{\natexlab{c}}.

\bibitem[Yuksekgonul et~al.(2022)Yuksekgonul, Bianchi, Kalluri, Jurafsky, and Zou]{yuksekgonul2022and}
Mert Yuksekgonul, Federico Bianchi, Pratyusha Kalluri, Dan Jurafsky, and James Zou.
\newblock When and why vision-language models behave like bags-of-words, and what to do about it?
\newblock \emph{arXiv preprint arXiv:2210.01936}, 2022.

\bibitem[Zang et~al.(2023)Zang, Li, Han, Zhou, and Loy]{zang2023contextual}
Yuhang Zang, Wei Li, Jun Han, Kaiyang Zhou, and Chen~Change Loy.
\newblock Contextual object detection with multimodal large language models, 2023.

\bibitem[Zhang et~al.(2021{\natexlab{a}})Zhang, Liao, Liu, Lu, Wang, Gao, and Li]{zhang2021CDN}
Aixi Zhang, Yue Liao, Si Liu, Miao Lu, Yongliang Wang, Chen Gao, and Xiaobo Li.
\newblock Mining the benefits of two-stage and one-stage hoi detection.
\newblock \emph{Advances in Neural Information Processing Systems}, 34:\penalty0 17209--17220, 2021{\natexlab{a}}.

\bibitem[Zhang et~al.(2021{\natexlab{b}})Zhang, Campbell, and Gould]{zhang2021scg}
Frederic~Z. Zhang, Dylan Campbell, and Stephen Gould.
\newblock Spatially conditioned graphs for detecting human–object interactions.
\newblock In \emph{Proceedings of the IEEE/CVF International Conference on Computer Vision (ICCV)}, pages 13319--13327, 2021{\natexlab{b}}.

\bibitem[Zhang et~al.(2022{\natexlab{a}})Zhang, Campbell, and Gould]{zhang2022UPT}
Frederic~Z Zhang, Dylan Campbell, and Stephen Gould.
\newblock Efficient two-stage detection of human-object interactions with a novel unary-pairwise transformer.
\newblock In \emph{Proceedings of the IEEE/CVF Conference on Computer Vision and Pattern Recognition}, pages 20104--20112, 2022{\natexlab{a}}.

\bibitem[Zhang et~al.(2023{\natexlab{a}})Zhang, Yuan, Campbell, Zhong, and Gould]{zhang2023pvic}
Frederic~Z. Zhang, Yuhui Yuan, Dylan Campbell, Zhuoyao Zhong, and Stephen Gould.
\newblock Exploring predicate visual context in detecting human–object interactions.
\newblock In \emph{Proceedings of the IEEE/CVF International Conference on Computer Vision (ICCV)}, pages 10411--10421, 2023{\natexlab{a}}.

\bibitem[Zhang et~al.(2023{\natexlab{b}})Zhang, Xie, and Deng]{zhang2023fine}
Rui Zhang, Cong Xie, and Liwei Deng.
\newblock A fine-grained object detection model for aerial images based on yolov5 deep neural network.
\newblock \emph{Chinese Journal of Electronics}, 32\penalty0 (1):\penalty0 51--63, 2023{\natexlab{b}}.

\bibitem[Zhang et~al.(2024)Zhang, Fu, and Zhang]{zhang2024deep}
Tao Zhang, Ying Fu, and Jun Zhang.
\newblock Deep guided attention network for joint denoising and demosaicing in real image.
\newblock \emph{Chinese Journal of Electronics}, 33\penalty0 (1):\penalty0 303--312, 2024.

\bibitem[Zhang et~al.(2022{\natexlab{b}})Zhang, Pan, Yao, Huang, Mei, and Chen]{Zhang_2022_STIP}
Yong Zhang, Yingwei Pan, Ting Yao, Rui Huang, Tao Mei, and Chang-Wen Chen.
\newblock Exploring structure-aware transformer over interaction proposals for human-object interaction detection.
\newblock In \emph{Proceedings of the IEEE/CVF Conference on Computer Vision and Pattern Recognition (CVPR)}, pages 19548--19557, 2022{\natexlab{b}}.

\bibitem[Zheng et~al.(2023{\natexlab{a}})Zheng, Gong, Jin, Peng, and Liu]{zheng2023generating}
Minghang Zheng, Shaogang Gong, Hailin Jin, Yuxin Peng, and Yang Liu.
\newblock Generating structured pseudo labels for noise-resistant zero-shot video sentence localization.
\newblock In \emph{Proceedings of the 61st Annual Meeting of the Association for Computational Linguistics (Volume 1: Long Papers)}, pages 14197--14209, 2023{\natexlab{a}}.

\bibitem[Zheng et~al.(2024)Zheng, Cai, Chen, Peng, and Liu]{zheng2024training}
Minghang Zheng, Xinhao Cai, Qingchao Chen, Yuxin Peng, and Yang Liu.
\newblock Training-free video temporal grounding using large-scale pre-trained models.
\newblock In \emph{European Conference on Computer Vision}, pages 20--37. Springer, 2024.

\bibitem[Zheng et~al.(2023{\natexlab{b}})Zheng, Xu, and Jin]{Zheng_2023_CVPR}
Sipeng Zheng, Boshen Xu, and Qin Jin.
\newblock Open-category human-object interaction pre-training via language modeling framework.
\newblock In \emph{Proceedings of the IEEE/CVF Conference on Computer Vision and Pattern Recognition (CVPR)}, pages 19392--19402, 2023{\natexlab{b}}.

\bibitem[Zhong et~al.(2020)Zhong, Ding, Qu, and Tao]{zhong2020polysemy}
Xubin Zhong, Changxing Ding, Xian Qu, and Dacheng Tao.
\newblock Polysemy deciphering network for human-object interaction detection.
\newblock In \emph{Computer Vision--ECCV 2020: 16th European Conference, Glasgow, UK, August 23--28, 2020, Proceedings, Part XX 16}, pages 69--85. Springer, 2020.

\bibitem[Zhong et~al.(2021)Zhong, Qu, Ding, and Tao]{zhong2021glance}
Xubin Zhong, Xian Qu, Changxing Ding, and Dacheng Tao.
\newblock Glance and gaze: Inferring action-aware points for one-stage human-object interaction detection.
\newblock In \emph{Proceedings of the IEEE/CVF Conference on Computer Vision and Pattern Recognition}, pages 13234--13243, 2021.

\bibitem[Zhou et~al.(2022)Zhou, Liu, Wang, Wang, Hu, Ding, and Wang]{zhou2022human}
Desen Zhou, Zhichao Liu, Jian Wang, Leshan Wang, Tao Hu, Errui Ding, and Jingdong Wang.
\newblock Human-object interaction detection via disentangled transformer.
\newblock In \emph{Proceedings of the IEEE/CVF conference on computer vision and pattern recognition}, pages 19568--19577, 2022.

\bibitem[Zhou and Chi(2019)]{zhou2019RLIP}
Penghao Zhou and Mingmin Chi.
\newblock Relation parsing neural network for human-object interaction detection.
\newblock In \emph{Proceedings of the IEEE/CVF International Conference on Computer Vision}, pages 843--851, 2019.

\end{thebibliography}
